\documentclass[10pt,twocolumn,letterpaper]{article}

\usepackage[pagenumbers]{cvpr} 

\usepackage[accsupp]{axessibility}
\usepackage{algorithm}
\usepackage{algpseudocodex}
\usepackage{amsmath}
\usepackage{amssymb}
\usepackage{multirow}
\usepackage{booktabs}
\usepackage{makecell}
\usepackage{subcaption}
\usepackage{fancybox}
\usepackage{fancyvrb}
\usepackage{bm}
\usepackage{multirow}
\usepackage{booktabs}
\usepackage{bbding}
\usepackage{makecell}
\usepackage{subcaption}
\usepackage{fancybox}
\usepackage{fancyvrb}

\definecolor{cvprblue}{rgb}{0.21,0.49,0.74}
\usepackage[pagebackref,breaklinks,colorlinks,citecolor=cvprblue]{hyperref}

\def\paperID{2485} 
\def\confName{CVPR}
\def\confYear{2024}

\newcommand{\speed}{199}

\title{LangSplat: 3D Language Gaussian Splatting}

\author{
  Minghan Qin$^{1,\ast}$, \quad Wanhua Li$^{2,\ast,}$\textsuperscript{\Envelope},   \quad Jiawei Zhou$^{1,\ast}$, \quad Haoqian Wang$^{1,}$\textsuperscript{\Envelope}, \quad Hanspeter Pfister$^{2}$\\
  $^1$Tsinghua University \;\;
  $^2$Harvard University \\
  \texttt{\small qmh21@mails.tsinghua.edu.cn, wanhua@seas.harvard.edu, zhoujw22@mails.tsinghua.edu.cn} \\
  \texttt{\small wanghaoqian@tsinghua.edu.cn, pfister@seas.harvard.edu} 
}

\begin{document}

\twocolumn[{%
\renewcommand\twocolumn[1][]{#1}%
\maketitle
\begin{center}
    \centering
    \captionsetup{type=figure}
    \includegraphics[width=1\textwidth]{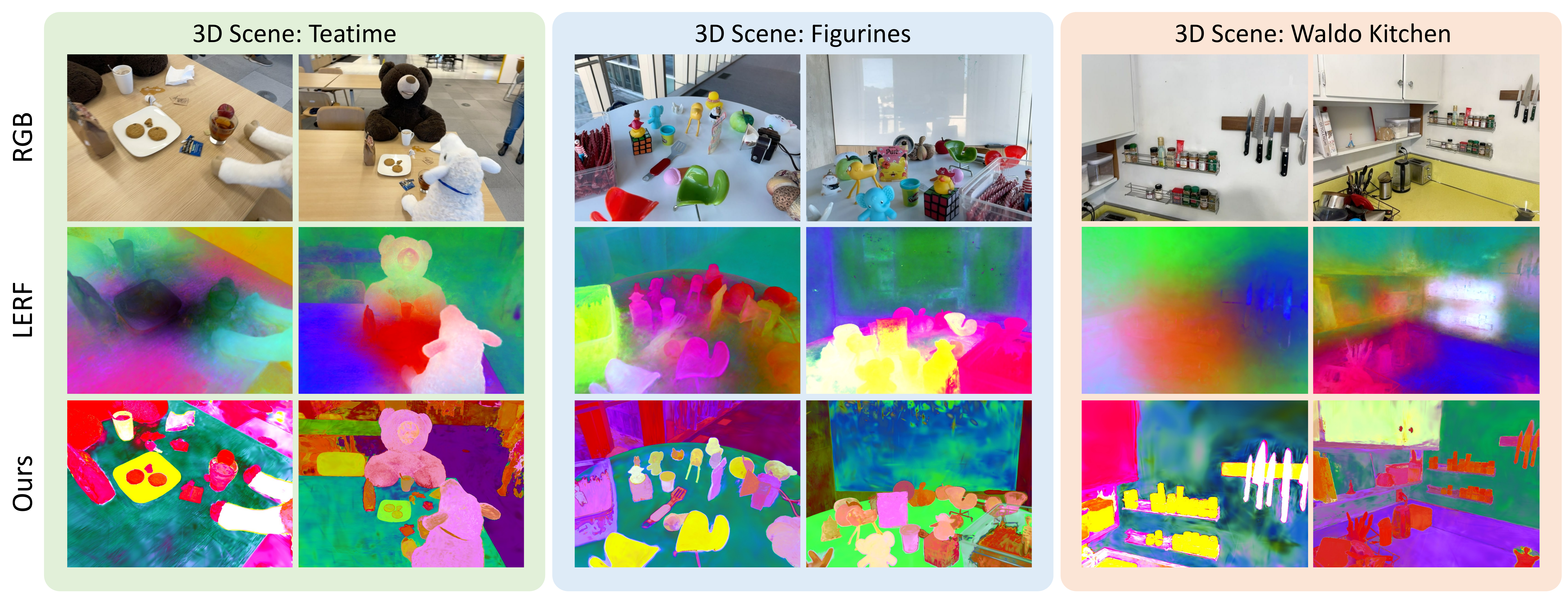}
    \captionof{figure}{Visualization of learned 3D language features of the previous SOTA method LERF and our LangSplat. While LERF generates imprecise and vague 3D features, our LangSplat accurately captures object boundaries and provides precise 3D language fields. While being effective, our LangSplat is also \textbf{{\speed}} $\times$ faster than LERF at the resolution of 1440 $\times$ 1080. 
    }
    \label{fig:teaser}
\vspace{-6pt}
\end{center}%
}]
{\let\thefootnote\relax\footnotetext{{$^{\ast}$ Equal contribution.\ \textsuperscript{\Envelope}Corresponding authors.}}}

\begin{abstract}
Humans live in a 3D world and commonly use natural language to interact with a 3D scene. Modeling a 3D language field to support open-ended language queries in 3D has gained increasing attention recently. This paper introduces LangSplat, which constructs a 3D language field that enables precise and efficient open-vocabulary querying within 3D spaces. Unlike existing methods that ground CLIP language embeddings in a NeRF model, LangSplat advances the field by utilizing a collection of 3D Gaussians, each encoding language features distilled from CLIP, to represent the language field. By employing a tile-based splatting technique for rendering language features, we circumvent the costly rendering process inherent in NeRF. Instead of directly learning CLIP embeddings, LangSplat first trains a scene-wise language autoencoder and then learns language features on the scene-specific latent space, thereby alleviating substantial memory demands imposed by explicit modeling. Existing methods struggle with imprecise and vague 3D language fields, which fail to discern clear boundaries between objects. We delve into this issue and propose to learn hierarchical semantics using SAM, thereby eliminating the need for extensively querying the language field across various scales and the regularization of DINO features. Extensive experimental results show that LangSplat significantly outperforms the previous state-of-the-art method LERF by a large margin. Notably, LangSplat is extremely efficient, achieving a {\speed} $\times$ speedup compared to LERF at the resolution of 1440 $\times$ 1080. We strongly recommend readers to check out our video results at \url{https://langsplat.github.io/}.
\end{abstract}

\section{Introduction}
\label{sec:intro}

Language is the primary means of communication for human beings~\cite{bonvillain2019language}. Modeling a 3D language field allows users to interact with and query 3D worlds using open-ended language, which presents a promising avenue for human-computer interaction and understanding~\cite{azuma2022scanqa,cascante2022simvqa,gordon2018iqa}. The field of open-ended language queries in 3D has attracted increasing attention due to its various applications such as robotic navigation~\cite{huang2023visual} and manipulation~\cite{shen2023distilled}, 3D semantic understanding~\cite{chen2023open,zhi2021place} and editing~\cite{kobayashi2022decomposing}, autonomous driving~\cite{jatavallabhula2023conceptfusion}, and augmented/virtual reality~\cite{liu2023weakly}.

Due to the absence of large-scale and diverse 3D scene data with language annotations, the current prevailing approach like LERF~\cite{kerr2023lerf} involves feature distillation from off-the-shelf vision-language models such as CLIP into a 3D scene. 
However, these methods~\cite{kerr2023lerf,liu2023weakly} suffer from significant limitations in both speed and accuracy, severely constraining their practical applicability. To address these two issues, we revisit two key aspects of 3D language field modeling: the 3D modeling approach that bridges the gap between 2D and 3D, and the rendering target, which determines what to learn for a 3D point.
For the 3D modeling technology, most methods utilize neural radiance fields (NeRFs) to represent a 3D scene, where volume rendering techniques are employed to accumulate 3D points along a ray into a single pixel. While NeRF has been demonstrated for its powerful 3D representation capabilities~\cite{park2021nerfies,pumarola2021d,yu2021pixelnerf,chan2021pi,shen2022learning,shen2023sd}, the volume rendering nature leads to its computationally expensive rendering speed~\cite{muller2022instant,chen2022tensorf,sun2022direct}, which imposes notable constraints on the potential applications within the NeRF-based language field. 

Regarding the rendering target, learning a CLIP embedding for a 3D point could be ambiguous as CLIP embeddings are aligned with images rather than pixels. Employing CLIP embeddings from a cropped patch also raises the point ambiguity issue, as the same 3D position can be associated with semantic concepts of varying scales. For instance, a point located on a bear's nose should yield high response values for three distinct textual queries: \textit{``bear's nose''}, \textit{``bear's head''}, and \textit{``bear''} given that this point contributes to all three hierarchical regions. To deal with this issue, current methods~\cite{kerr2023lerf,liu2023weakly} introduce an additional absolute scale input to NeRF, trains with patch-wise CLIP features at different scales, and densely renders 2D maps at multiple scales during querying to select the optimal one. 
However, this scale-based solution compromises both efficiency and effectiveness. It could increase query time by up to 30 times as it needs to render at multiple different scales. Moreover, most patches with varying scales often fail to accurately encompass objects, either frequently including other objects from the background or omitting portions of the target object. These inaccurate CLIP features lead to the trained 3D language field lacking clear boundaries and containing a significant amount of noise. Therefore, they often simultaneously learn pixel-aligned DINO features to mitigate this issue. However, the performance remains unsatisfactory. As shown in Figure \ref{fig:teaser}, LERF still generates imprecise 3D language features.

In this paper, we propose the 3D Language Gaussian Splatting (LangSplat) to address the above issues. Instead of using NeRF to build 3D representations, we resort to 3D Gaussian Splatting,
which represents a 3D scene as a collection of 3D Gaussians and uses tile-based splatting to achieve efficient rendering at high resolutions. Our LangSplat defines a set of 3D language Gaussians, with each Gaussian being enhanced by a language embedding. These language-enhanced Gaussians are supervised using CLIP embeddings extracted from image patches captured from multiple training views, ensuring multi-view consistency.  As an explicit modeling method, directly storing the high-dimensional language embeddings for each 3D language Gaussian is memory-inefficient. To reduce the memory cost and further improve the rendering efficiency, we propose to first learn a scene-wise language autoencoder, which maps CLIP embeddings in a scene to a low-dimensional latent space. In this way, each language Gaussian only contains the low-dimensional latent language features and the final language embeddings are obtained with decoding of the rendered features. To address the point ambiguity issue, we propose to employ the semantic hierarchy defined by the Segment Anything Model (SAM)~\cite{kirillov2023segment}. Specifically, for each 2D image,  we obtain three well-segmented maps at different semantic levels with SAM. Then we extract the CLIP feature for each mask with precise object boundaries and assign this feature to every point on the corresponding mask. Learning with SAM-based masks not only endows each point with precise CLIP embeddings, resulting in higher model accuracy, but also enables direct querying at predefined three semantic scales. This circumvents the need for intensive searches across multiple absolute scales and the auxiliary DINO features, thereby effectively improving efficiency.
We summarize the contributions of this paper as follows:

\begin{itemize}
\item We propose the LangSplat, which is the first 3D Gaussian Splatting-based method for 3D language fields. A scene-specific autoencoder is further introduced to alleviate the memory cost issue imposed by explicit modeling.

\item We propose to learn the hierarchical semantics defined by SAM to address the point ambiguity issue for 3D language field modeling.

\item Experimental results show that our method outperforms the state-of-the-art methods on open-vocabulary 3D object localization and semantic segmentation tasks while being {\speed} $\times$ faster than LERF at 1440 $\times$ 1080 resolution.
\end{itemize}

\section{Related Work}
\label{sec:related}

\textbf{3D Gaussian Splatting.} Real-time rendering has always been a pursued objective for neural rendering. Recently, Kerbl \etal~\cite{kerbl20233d} proposed to represent the 3D scene with a set of 3D Gaussians, which attained real-time rendering for 1080p resolution while maintaining state-of-the-art visual quality. Encouraged by the success of 3D Gaussian Splatting, many works extend it to other tasks.
To achieve real-time dynamic scene rendering, some studies~\cite{wu20234d,luiten2023dynamic,yang2023real} have extended the 3D Gaussian Splatting technique to dynamic scenes. Luiten \etal~\cite{luiten2023dynamic} proposed the Dynamic 3D Gaussians, which extended 3D Gaussians to dynamic scenes by explicitly modeling the 3D Guassians across different time steps. Yang \etal~\cite{yang2023deformable} presented a deformable 3D Gaussians Splatting method, which learned 3D Gaussians in canonical space and modeled the dynamic scenes with a deformation field. Meanwhile, some researchers have combined 3D Gaussian Splatting with diffusion models to achieve efficient text-to-3D generation~\cite{tang2023dreamgaussian,yi2023gaussiandreamer,chen2023text}. For example, Tang \etal~\cite{tang2023dreamgaussian} introduced DreamGaussian for efficient 3D content generation with a generative 3D Gaussian Splatting model. Unlike these methods, our paper extends each 3D Gaussian with language embeddings for open-vocabulary 3D queries.

\noindent\textbf{SAM.} The Segment Anything Model~\cite{kirillov2023segment}, released by Meta in 2023, has attracted considerable attention~\cite{huang2023instruct2act,mai2023opportunities,ma2023segment}, 
which achieved impressive zero-shot performance.
It has become the foundational model for image segmentation. 
SAM has been used for many vision tasks~\cite{yao2023matte,mazurowski2023segment,lu2023can,wan2024trisam} such as image inpainting~\cite{yu2023inpaint}, object tracking~\cite{cheng2023segment,yang2023track}, image editing~\cite{gao2023editanything}, and so on. 
Many efforts have also been made to utilize SAM in the 3D domain~\cite{shen2023anything,cen2023segment}. Liu \etal~\cite{liu2023segment} proposed Seal to explore the potential of VFMs including SAM for point cloud segmentation.  SA3D~\cite{cen2023segment} generalized SAM to 3D objects by leveraging NeRF to connect 2D images and 3D space.  
Different from these works, we use SAM to obtain accurate object masks with three well-defined hierarchical semantics to train a 3D language field.

\noindent\textbf{3D Langugae Fields.} Some early attempts to construct 3D feature fields included Distilled Feature Fields~\cite{kobayashi2022decomposing} and Neural Feature Fusion Fields~\cite{tschernezki2022neural}. They learned 3D-consistent features by distilling LSeg~\cite{li2022languagedriven} or DINO~\cite{caron2021emerging} features across multiple views into a NeRF. Shen \etal~\cite{shen2023distilled} further adopted distilled feature fields for few-shot language-guided robotic manipulation by distilling CLIP feature into a NeRF.
There are also some efforts~\cite{zhi2021place,siddiqui2023panoptic} that embed semantic information into NeRFs. For example, Semantic NeRF~\cite{zhi2021place} jointly encoded semantics with appearance and geometry within a NeRF for novel semantic view synthesis. 
LERF~\cite{kerr2023lerf} was the first to embed CLIP features into NeRF, enabling open-vocabulary 3D queries leveraging the powerful CLIP representation. DINO features were also used for supervising LERF to improve its performance. Liu \etal~\cite{liu2023weakly} also utilized CLIP and DINO features to train a NeRF model for 3D open-vocabulary segmentation. While these methods use NeRF for 3D modeling and suffer from the costly rendering process, we propose the 3D language Gaussian Splatting to obtain efficient 3D language fields.

\section{Proposed Approach}
\label{sec:method}
\begin{figure*}[ht!]
    \centering
    \includegraphics[width=1.0\textwidth]{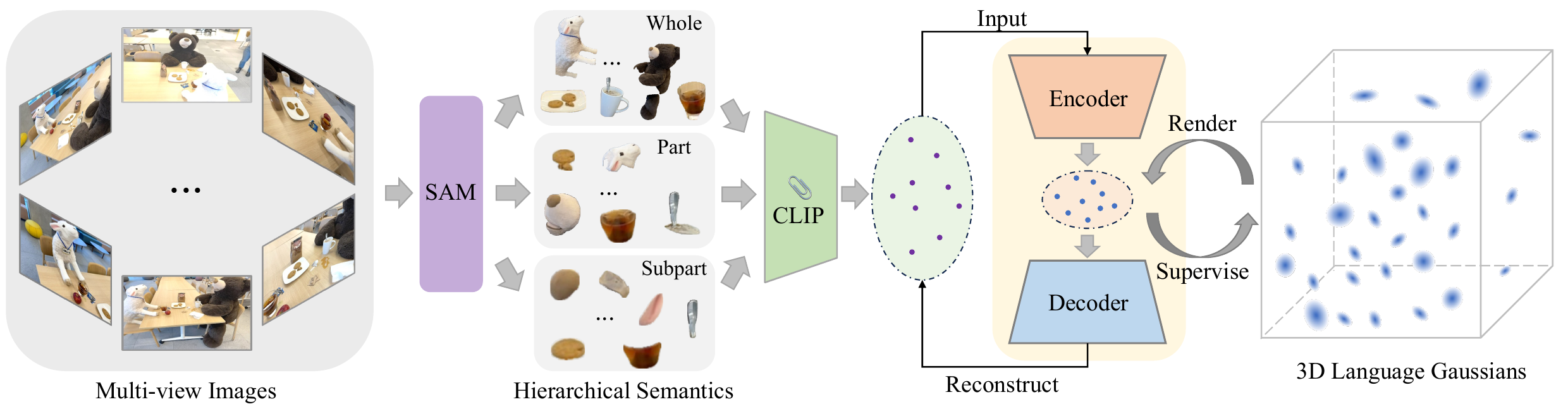}
    \caption{ The framework of our LangSplat. Our LangSplat leverages SAM to learn hierarchical semantics to address the point ambiguity issue. Then segment masks are sent to the CLIP image encoder to extract the corresponding CLIP embeddings. We learn an autoencoder with these obtained CLIP embeddings. Our 3D language Gaussian learn language features on the scene-specific latent space to reduce the memory cost. During querying, the rendered language embeddings are sent to the decoder to recover the features on the CLIP space.
    }  \label{fig:framework}
    \vspace{-14pt}
\end{figure*}

In this section, we first revisit the challenges of modeling 3D language fields and then elaborate on how our proposed LangSplat addresses these issues. Figure \ref{fig:framework} depicts the framework of our proposed LangSplat.

\subsection{Revisiting the Challenges of Language Fields}

We denote an input image as $\bm{I} \in \mathbb{R}^{3 \times H \times W}$, where $H$ and $W$ represent the height and weight of the image size. We take a set of calibrated images $\{ \bm{I}_t | t =1,2,...T\}$ as input and train a 3D language field $\Phi$ with these images. Most existing methods~\cite{kerr2023lerf,liu2023weakly,shafiullah2022clip} employ the CLIP image encoder $V$ to extract image features and utilize the extracted CLIP embeddings to supervise the 3D language field $\Phi$, leveraging the well-aligned text-image latent space~\cite{radford2021learning,li2022ordinalclip,zhou2022learning} provided by CLIP, thus facilitating open-vocabulary queries. However, CLIP embeddings are image-aligned rather than pixel-aligned. In other words, simply computing $V(\bm{I}_t) \in \mathbb{R}^D$ only obtains an image-level feature, whereas what we need is a pixel-aligned language embedding $\bm{L}_t \in \mathbb{R}^ {D \times H \times W}$, where $D$ represents the CLIP feature dimension. Meanwhile, modeling pixel-aligned language features faces the issue of point ambiguity, as a single point on an object contributes to multiple semantic levels of regions. For instance, a point on a cat's ear simultaneously contributes to the cat's ear, the cat's head, and the entire cat, and should be activated by all three types of textual queries.

To address these issues, most existing methods~\cite{kerr2023lerf,liu2023weakly} extract a hierarchy of CLIP features from cropped image patches. Specifically, for a pixel with coordinates $v \in \{1, ..., H\} \times \{1, ..., W\}$, the corresponding CLIP features are obtained from image patches centered around $v$ at different absolute physical scales $s$, with the expectation that at a certain scale $s$, the patch can fully encompass the object. However, this multi-scale approach has two limitations. Firstly, patch features are imprecise because they often include additional contextual object information, leading to overly smoothed language fields with indistinct object boundaries. To alleviate the patchy issue, most methods~\cite{kerr2023lerf,liu2023weakly} leverage additional pixel-aligned DINO features to supervise the network. However, the learned 3D language features are still imprecise, as illustrated in Figure \ref{fig:teaser}. Secondly, it requires simultaneous rendering at multiple scales during inference to find the optimal scale. With the number of scales $s$ potentially reaching as high as 30~\cite{kerr2023lerf}, this significantly diminishes the inference speed. 

Besides the rendering target, another key design space is the 3D modeling approach. Most existing methods~\cite{tschernezki2022neural,bhalgat2023contrastive} employ NeRFs for 3D representation, where they learn a language feature at each 3D point and subsequently render the language feature onto an image, similar to color rendering. However, NeRF-based methods are constrained by their time-consuming rendering process, even though the most advanced NeRF techniques currently available cannot achieve real-time rendering in high-resolution, unrestricted scenes~\cite{kerbl20233d}. Meanwhile, there is a high demand for efficient open vocabulary querying in practical applications, especially in fields such as intelligent robotics.

\subsection{Learning Hierarchical Semantics with SAM}

As a foundation model for image segmentation, SAM~\cite{kirillov2023segment} can accurately group a pixel with its surrounding pixels belonging to the same object, thereby segmenting the image into many object masks with clear boundaries. Furthermore, SAM addresses point ambiguity by generating three different masks for a point prompt, namely, \textit{whole}, \textit{part}, and \textit{subpart}, representing three hierarchical levels of semantics.
In this paper, we propose leveraging SAM to obtain precise object masks, which are then used to acquire pixel-aligned features. We also explicitly model the semantic hierarchy defined by SAM to address the point ambiguity issue. With SAM, we can capture the semantic hierarchy of objects in 3D scenes, providing accurate and multi-scale segmentation maps for each input image.

Specifically, we feed a regular grid of $32 \times 32$ point prompts into SAM to obtain the masks under three different semantic levels: $\bm{M}_0^s, \bm{M}_0^p, \bm{M}_0^w$, where $\bm{M}_0^s, \bm{M}_0^p$, and $\bm{M}_0^w$ represent the masks at subpart, part, and whole levels, respectively. Then we remove redundant masks for each of the three mask sets based on the predicted IoU score, stability score, and overlap rate between masks. Each filtered mask set independently performs a comprehensive full-image segmentation based on its respective semantic level, resulting in three segmentation maps: $\bm{M}^s, \bm{M}^p, \bm{M}^w$. These segmentation maps precisely delineate the boundaries of objects at their hierarchical levels, effectively partitioning the scene into semantically meaningful regions. With the obtained segmentation maps, we proceed to extract CLIP features for each segmented region. These features capture the semantic context of the objects at various levels within the scene. Mathematically, the obtained pixel-aligned language embeddings are:
\begin{equation}
    \bm{L}^l_t(v) = V (\bm{I}_t \odot \bm{M}^l(v)), l \in \{ s,p,w \},
    \label{eq:samtarget}
\end{equation}
where $\bm{M}^l(v)$ represents the mask region to which pixel $v$ belongs at the semantic level $l$.

Each pixel rendered from the 3D language scene now possesses a CLIP feature that aligns with its precise semantic context. This alignment reduces ambiguity and enhances the accuracy of language-based queries.  We can learn an accurate 3D language field even without the commonly used DINO regularization. 
Another advantage of our SAM-based approach is the predefined semantic scales. Since we have distinct segmentation maps for ``whole," ``part," and ``subpart" levels, we can directly query the 3D language field at these predefined scales. This eliminates the need for intensive searches across multiple absolute scales, making the querying process more efficient.
By incorporating SAM's semantic hierarchy into our approach, we not only improve the accuracy of our 3D language field but also streamline the querying process, making it more efficient and effective for a wide range of applications.

\subsection{3D Gaussian Splatting for Language Fields}

Having obtained the language embeddings on a set of 2D images $\{\bm{L}^l_t, | t =1, ..., T\}$, we can learn a 3D language scene by modeling the relations between 3D points and 2D pixels.  Most existing methods~\cite{tschernezki2022neural,bhalgat2023contrastive} suffer from the costly rendering process as they adopt NeRFs for 3D modeling. To address this issue, we present the first 3D Gaussian Splatting-based method for 3D language field modeling. 

3D Gaussian Splatting explicitly represents a 3D scene as a collection of anisotropic 3D Gaussians, with each Gaussian $G(x)$ characterized by 
 a mean $\mu \in \mathbb{R}^3$ and a covariance matrix $\Sigma$:
\begin{equation}
    G(x) = \exp (- \frac{1}{2} (x - \mu)^{\top} \Sigma^{-1} (x -  \mu)).
    \label{eq:3dGaussian}
\end{equation}

To optimize the parameters of 3D Gaussians, they are rendered into 2D image planes~\cite{zwicker2001ewa}, and a tile-based rasterizer is used to improve the rendering efficiency:
\begin{equation}
    C(v) = \sum_{i \in \mathcal{N}} c_i \alpha_i \prod_{j=1}^{i-1} (1 - \alpha_j),
    \label{eq:rendering_3dgs}
\end{equation}
where $c_i$ is the color of the $i$-th Gaussian, $\mathcal{N}$ denotes the Gaussians in the tile, $C(v)$ is the rendered color at pixel $v$, and $\alpha_i = o_i G^{2D}_i(v)$. Here $o_i$ is the opacity of the $i$-th Gaussian and $G^{2D}_i(\cdot)$ represents the function of the $i$-th Gaussian projected onto 2D.

In this paper, we proposes the 3D language Gaussian Splatting, which augments each 3D Gaussian with three language embeddings $\{\bm{f}^s,\bm{f}^p,\bm{f}^w\}$. These embeddings are derived from CLIP features, which capture the hierarchical semantics provided by SAM. The augmented Gaussians are named as 3D language Gaussians. We also adopt the tile-based rasterizer to retain the rendering efficiency: 
\begin{equation}
    \bm{F}^l(v) = \sum_{i \in \mathcal{N}} \bm{f}_i^l \alpha_i \prod_{j=1}^{i-1} (1 - \alpha_j), l \in \{s,p,w\},
    \label{eq:rendering_lang}
\end{equation}
where $\bm{F}^l(v)$ represents the language embedding rendered at pixel $v$ with the semantic level $l$. By incorporating language information directly into the Gaussians, we enable the 3D language field to respond to language-based queries.

As an explicit modeling approach, our LangSplat may create millions of 3D points to model a complex 3D scene. As CLIP embeddings are high-dimensional features, directly learning $\bm{f}^l$ on the CLIP latent space significantly increases 
memory and time cost.
Compared to learning RGB colors without spherical harmonics coefficients, learning 512-dimensional CLIP features increases the memory requirements for storing 3D Gaussians by over 35 times, 
easily leading to running out of L1 cache memory.
To reduce memory cost and improve efficiency, we introduce a scene-wise language autoencoder. This autoencoder maps CLIP embeddings in a scene to a lower-dimensional latent space, reducing memory requirements. The CLIP model is trained using 400 million (image, text) pairs, and its $D$-dimensional latent space could be highly compact, as it needs to align arbitrary text and images in this space. However, the language field $\Phi$ we train here is scene-specific, meaning we can leverage scene priors to compress CLIP features. In fact, for each input image, we will obtain hundreds of masks segmented by SAM, which is significantly smaller than the number of images used in CLIP training. Therefore, all the segmented regions in a scene are sparsely distributed in the CLIP latent space, allowing us to further compress these CLIP features using a scene-specific autoencoder.

Specifically, we use the collections of CLIP features of SAM segmented masks $\{\bm{L}_t^l | l \in \{s,p,w\}, 1 \leq t \leq T\}$ to train a lightweight autoencoder. An encoder $E$ maps the $D$-dimensional CLIP features $\bm{L}_t^l(v) \in \mathbb{R}^D$ to $\bm{H}_t^l(v) = E(\bm{L}_t^l(v)) \in \mathbb{R}^d$, where $d \ll D$. Then we learn a decoder $\Psi$ to reconstruct the original CLIP embeddings from the compressed representation. The autoencoder is trained with a reconstruction objective on the CLIP embeddings $\{\bm{L}_t^l\}$:
\begin{equation}
     \mathcal{L}_{ae}=  \sum_{l \in \{s,p,w\}} \sum_{t=1}^{T} d_{ae}(\Psi(E(\bm{L}_t^l(v))), \bm{L}_t^l(v)),
    \label{eq:loss_ae}
\end{equation}
where $d_{ae}()$ denotes a distance function used for the autoencoder. Here we adopt both $\mathcal{L}_1$ and a cosine distance loss. 

After training the autoencoder, we transform all CLIP embeddings $\{\bm{L}_t^l\}$ into scene-specific latent features $\{\bm{H}_t^l\}$. We let our 3D language Gaussians learn language embeddings in the scene-specific latent space instead of the CLIP latent space. Therefore, we have $\bm{f}^l \in \mathbb{R}^d$. In practice, we choose $d =3$ as it yields excellent model efficiency and accuracy. Compared to directly modeling the $D$-dimensional CLIP embeddings, our method significantly reduced the memory cost by incorporating scene priors. We optimized the language embeddings with the objective:
\begin{equation}
     \mathcal{L}_{lang}=  \sum_{l \in \{s,p,w\}} \sum_{t=1}^{T} d_{lang}( \bm{F}_t^l(v), \bm{H}_t^l(v)),
    \label{eq:loss_langsplat}
\end{equation}
where $d_{lang}()$ denotes the distance function used for our 3D Language Gaussians.

During inference, we follow the Eq. \eqref{eq:rendering_lang} to render the language embeddings from 3D to 2D, and then we use the trained scene-specific decoder $\Psi$ to recover the CLIP image embeddings $\Psi(\bm{F_t^l}) \in \mathbb{R}^{D \times H \times W}$, which enable open-vocabulary queries with the CLIP text encoder.

By enhancing 3D Gaussians with language embedding and employing a scene-wise language autoencoder, our proposed LangSplat presents a powerful and efficient solution for building 3D language fields. This approach not only preserves the rendering efficiency of Gaussian Splatting but also mitigates the catastrophic memory explosion associated with explicit modeling. 

\subsection{Open-vocabulary Querying}

Due to the well-aligned latent space between images and text provided by the CLIP model, our learned 3D language field can easily support open-vocabulary 3D queries, including open-vocabulary 3D object localization and open-vocabulary 3D semantic segmentation. 
Many existing open-vocabulary 3D semantic segmentation methods~\cite{liu2023weakly} usually select the category from a category list, which includes the categories present in the images. However, obtaining a comprehensive category list for in-the-wild scenes is challenging. Different from them, our method generates precise object masks given an arbitrary text query.

Following LERF~\cite{kerr2023lerf}, we compute the relevancy score for each text query. Specifically, for each rendered language embedding $\phi_{img}$ and each text query $\phi_{qry}$, the relevancy score is defined as  $\min_i \frac{\exp(\phi_{img} \cdot \phi_{qry})}{\exp(\phi_{img} \cdot \phi_{qry}) + \exp(\phi_{img} \cdot \phi_{canon}^i)}$, where $\phi_{canon}^i$ is the CLIP embeddings of a predefined canonical phrase chosen from \textit{``object"}, \textit{``things"}, \textit{``stuff"}, and \textit{``texture"}. Hence, for each text query, we will obtain three relevancy maps, each representing results at a specific semantic level. We follow the strategy used in LERF~\cite{kerr2023lerf} and choose the semantic level that yields the highest relevancy score. For the 3D object localization task, we directly choose the point with the highest relevance score. For the 3D semantic segmentation task, we filter out points with relevancy scores lower than a chosen threshold, and predict the object masks with remaining regions. Please refer to the appendix for additional details.

    \section{Experiments}
\label{sec:experiment}

\subsection{Settings}

\textbf{Datasets.} We employ two datasets for evaluation. The LERF dataset~\cite{kerr2023lerf} is captured using the iPhone App Polycam, which consists of complex in-the-wild scenes. The LERF dataset is designed for 3D object localization tasks, here we extend the LERF dataset by annotating ground truth masks for textual queries, enabling the evaluation of the open-vocabulary 3D semantic segmentation on the LERF dataset. As the original LERF annotations for 3D object localization are relatively simple, performance in some scenarios has already approached saturation. Therefore, we further manually annotated additional challenging localization samples to better evaluate method performance. We report localization accuracy for the 3D object localization task following LERF~\cite{kerr2023lerf}, and report the IoU results for the 3D semantic segmentation task. We also employ the 3D-OVS dataset~\cite{liu2023weakly}, which comprises a collection of long-tail objects captured in diverse poses and backgrounds. This dataset is developed for open-vocabulary 3D semantic segmentation,  where a full list of categories is provided. While other methods use the full list to generate the predicted masks, we only use the query category to generate the corresponding masks. The mIoU metric is used for this dataset.

\noindent\textbf{Implementation Details.}
To extract the language features of each image, we utilize the OpenCLIP ViT-B/16 model. For SAM, we use the ViT-H model to segment 2D masks. For each scene, we first use 3D Gaussian Splatting to train an RGB scene. 
We train it for 30,000 iterations, and in the end, each scene comprises around 2,500,000 points.
We follow the default parameter setting as in~\cite{kerbl20233d} to train the RGB scene.
Then we train our 3D language Gaussians by fixing all other parameters of 3D Gaussians such as mean and opacity. Only the language features are learnable during this stage. 
We train the language features for 30,000 iterations.
Our autoencoder is implemented by an MLP, which compresses the 512-dimensional CLIP features into 3-dimensional latent features. 
For a $1440 \times 1080$ resolution scene,
our model is trained for $\sim$25 minutes on an NVIDIA RTX-3090 GPU 
and takes roughly 4GB of memory.

\begin{table}[t]
  \centering
  \begin{tabular}{lccc}
    \toprule
    Test Scene & LSeg~\cite{li2022languagedriven} & LERF~\cite{kerr2023lerf} & LangSplat \\
    \midrule
    ramen & 14.1  & 62.0 &   \textbf{73.2} \\
    figurines & 8.9 & 75.0  & \textbf{80.4} \\
    teatime & 33.9  & 84.8   & \textbf{88.1} \\
    waldo\_kitchen & 27.3 & 72.7 & \textbf{95.5} \\
    \midrule
    overall & 21.1  & 73.6 & \textbf{84.3} \\
    \bottomrule
  \end{tabular}
  \caption{Localization accuracy (\%) comparisons on LERF dataset.}
  \label{table:lerf_loca}
  \vspace{-5pt}
\end{table}

\begin{table}[t]
  \centering
  \begin{tabular}{lccc}
    \toprule
    Test Scene & LSeg~\cite{li2022languagedriven} & LERF~\cite{kerr2023lerf} & LangSplat \\
    \midrule
    ramen & 7.0  & 28.2 &   \textbf{51.2} \\
    figurines & 7.6 & 38.6  & \textbf{44.7} \\
    teatime & 21.7   & 45.0   & \textbf{65.1} \\
    waldo\_kitchen & 29.9 & 37.9 & \textbf{44.5} \\
    \midrule
    overall & 16.6  & 37.4 & \textbf{51.4} \\
    \bottomrule
  \end{tabular}
  \caption{Quantitative comparisons of 3D semantic segmentation on the LERF dataset. We report the average IoU scores (\%).}
  \label{table:lerf_seg}
  \vspace{-5pt}
\end{table}

\begin{figure*}[t]
  \centering
   \includegraphics[width=1\linewidth]{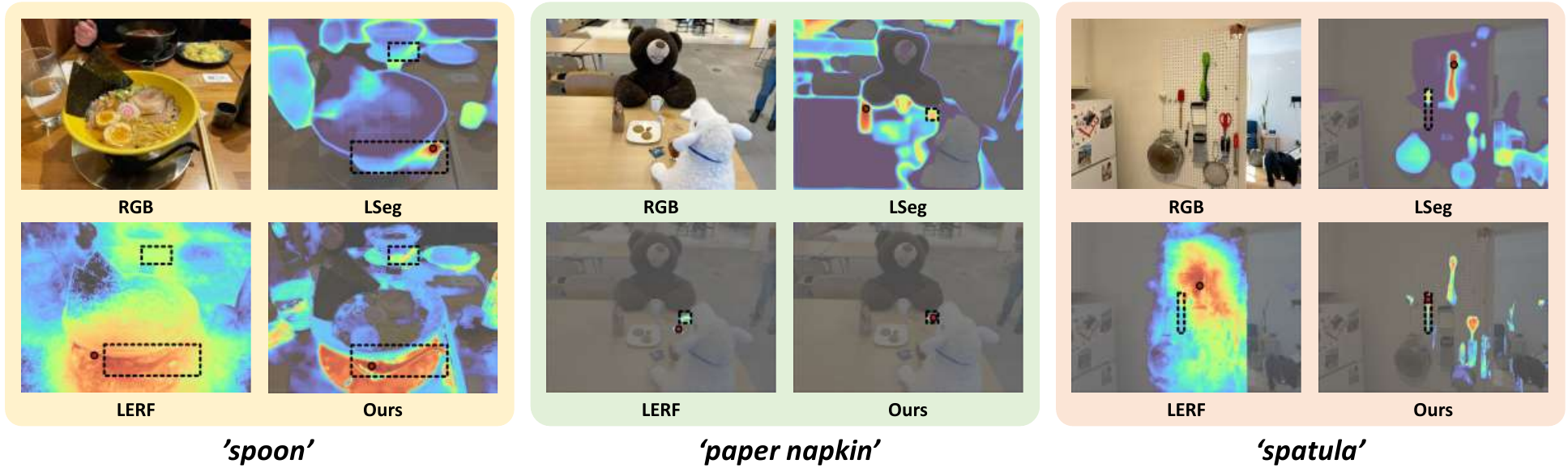}
   \caption{Qualitative comparisons of open-vocabulary 3D object localization on the LERF dataset. The red points are the model predictions and the black dashed bounding boxes denote the annotations.}
   \label{fig:localization}
    \vspace{-4pt}
\end{figure*}

\begin{figure*}[t]
  \centering
   \includegraphics[width=1\linewidth]{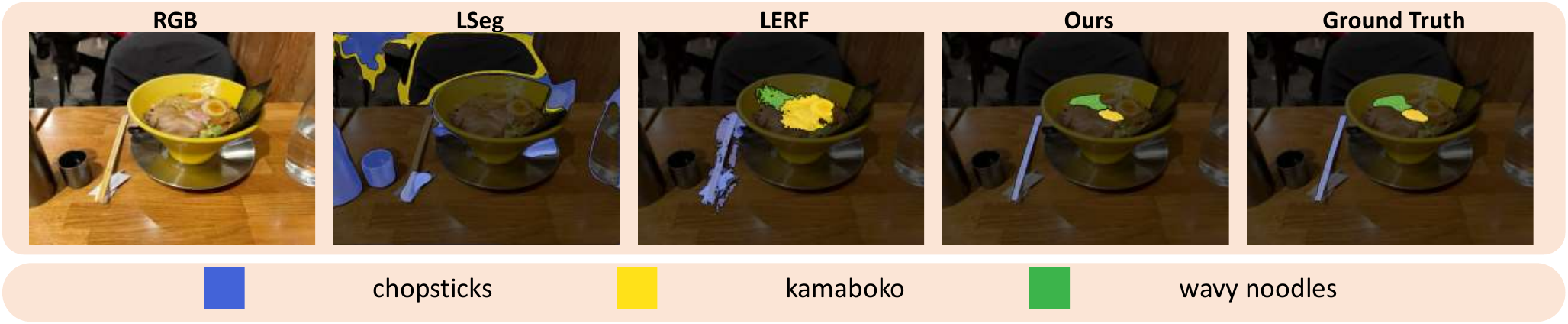}
   \caption{Qualitative comparisons of open-vocabulary 3D semantic segmentation on the LERF dataset.}
   \label{fig:LERFseg}
   \vspace{-12pt}
\end{figure*}

\subsection{Results on the LERF dataset}

\textbf{Quantitative Results.} We first compare our method with other methods on the LERF dataset. Table~\ref{table:lerf_loca} shows the localization results. We observe that our method achieves an overall accuracy of 84.3\%, significantly outperforming LERF. Table~\ref{table:lerf_seg} further shows the IoU results of 3D semantic segmentation, our method outperforms LERF by 14.0\%, which illustrates the superiority of our proposed LangSplat.

\noindent\textbf{Visualization Results.} To show the learned 3D language field, we visualize the learned features by computing 3-dimensional PCA components of learned language features following~\cite{kobayashi2022decomposing}. The results are shown in Figure~\ref{fig:teaser}. We see that the LERF learned features fail to generate clear boundaries between objects while our method gives precise object shapes solely using CLIP features.
We further show the visualization results of object localization and semantic segmentation in Figure~\ref{fig:localization} and Figure~\ref{fig:LERFseg}, respectively. We observe that the activation regions generated by LERF are more dispersed, while ours are more concentrated, and our activation regions can better align with the ground truth shape compared to those produced by LERF.

\begin{table}[t]
  \renewcommand\tabcolsep{8pt}
  \centering
  \begin{tabular}{ccc|cc}
    \toprule
    \multicolumn{3}{c|}{Component} & \multicolumn{2}{c}{Performance} \\
    \midrule
    AE & 3D-GS & SAM & IoU (\%) & Speed (s/q)  \\
    \midrule
     & & & 28.20 & 30.93\\
       &  & \Checkmark & 46.74 & 7.77 \\
         & \Checkmark & \Checkmark & OOM & OOM \\
    \Checkmark & \Checkmark & \Checkmark & \textbf{51.15} & \textbf{0.26} \\
    \bottomrule
  \end{tabular}
  \caption{Ablations. The results are obtained on the ramen scene.}
  \label{table:ablation}
  \vspace{-8pt}
\end{table}

\begin{table}[t]
  \renewcommand\tabcolsep{8pt}
  \centering
  \begin{tabular}{ccc|cc}
    \toprule
    \multicolumn{3}{c|}{Component} & \multicolumn{2}{c}{Performance} \\
    \midrule
    AE & 3D-GS & SAM & mIoU (\%)  & Speed (s/q)  \\
    \midrule
     & & & 53.2 & 55.7 \\
       &  & \Checkmark & 85.7 & 18.4 \\
         & \Checkmark & \Checkmark & OOM  & OOM \\
    \Checkmark & \Checkmark & \Checkmark & \textbf{94.2} & \textbf{0.28} \\
    \bottomrule
  \end{tabular}
  \caption{Ablations result on the bench scene of the 3D-OVS dataset. The image resolution is $1440 \times 1080$.}
  \label{table:ablation3dovs}
  \vspace{-8pt}
\end{table}

\noindent\textbf{Ablation Study.} We conduct ablations on the ramen scene and report the semantic segmentation results in Table \ref{table:ablation}. We test the query speed on an NVIDIA RTX-3090 GPU. Here AE represents autoencoder and 3D-GS denotes 3D Gaussian Splatting. Without any of our proposed components, our baseline equals LERF, which has a speed of 30.93 seconds per text query at the resolution of 988 $\times$ 731. Using SAM to replace the scale-based solution significantly increases the IoU by 18.54\%, showing our SAM-based solution effectively addresses the point ambiguity issue, leading to accurate 3D language features. Simply replacing NeRF with 3D Gaussian Splatting 
leads to running out of L1 cache memory 
as explicitly modeling CLIP features poses huge memory demands. Incorporating a scene-specific autoencoder effectively addresses this issue and results in further improvements in both accuracy and efficiency. In the end, our LangSplat achieved a 119 $\times$ speedup over LERF while significantly surpassing LERF in terms of accuracy.

\begin{figure*}[t]
  \centering
   \includegraphics[width=1\linewidth]{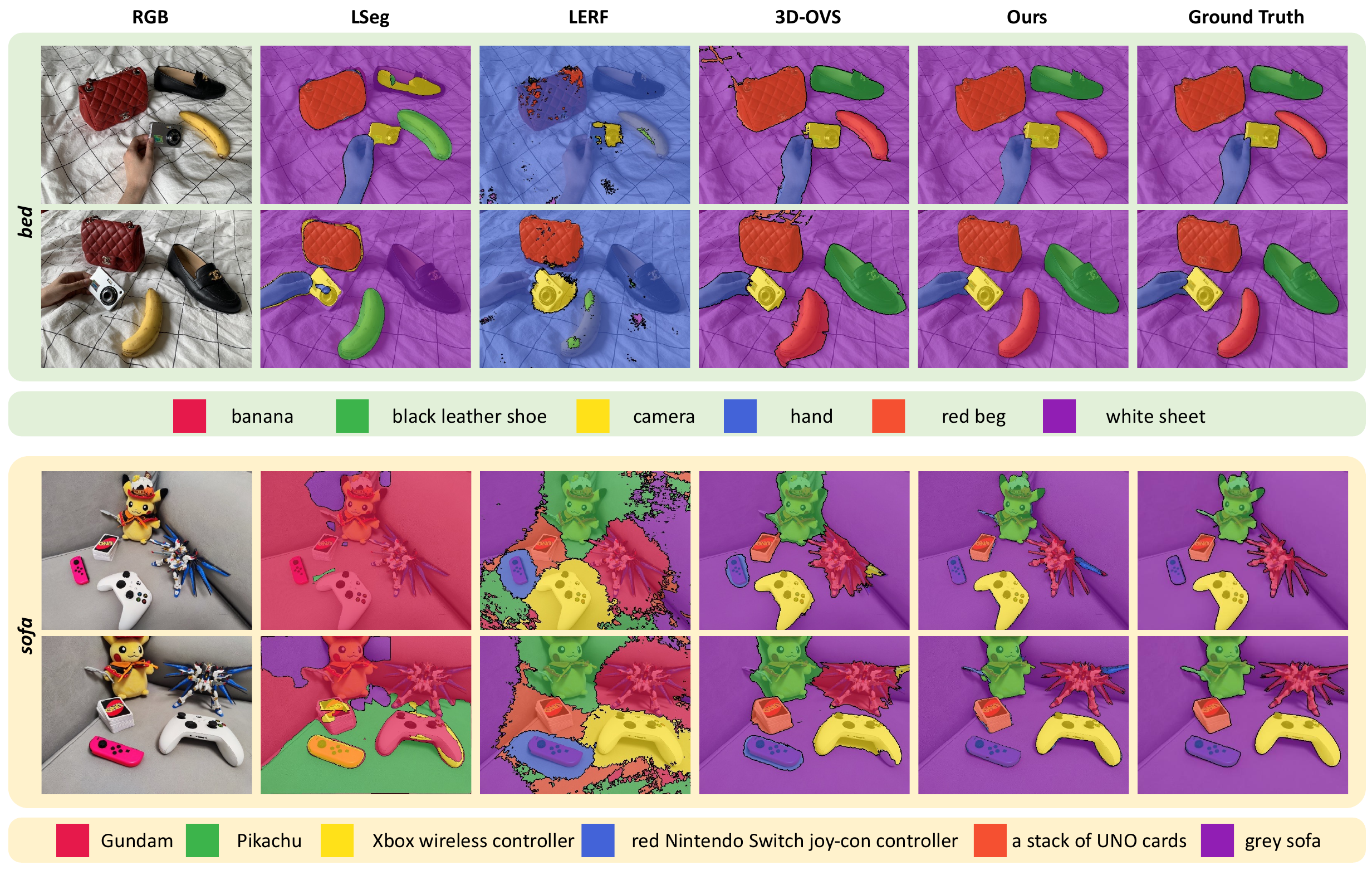}
   \caption{Qualitative comparisons of different methods on the 3D-OVS dataset. We visualize the segmentation results in 2 scenes. We observe that our method gives the most accurate segmentation maps.}
   \label{fig:3dovs}
   \vspace{-14pt}
\end{figure*}

We further conducted the ablations on the 3D-OVS dataset, which has a higher image resolution of $1440 \times 1080$. Table \ref{table:ablation3dovs} lists the results on the bench scene. We also tested the query speed on an NVIDIA RTX-3090 GPU. We observed that with the increase in image resolution, the speedup over LERF further improved to 199 $\times$, which demonstrates the huge potential of our method. Our further study shows that most of the computational time is allocated to the decoder rather than the rendering process. We could replace the decoder with a $1 \times 1$ convolutional layer to attain a higher speedup.

\begin{table}[t]
  \renewcommand\tabcolsep{4pt}
  \centering
  \begin{tabular}{lcccccc}
    \toprule
    Method & \textit{bed}   & \textit{bench} & \textit{room}  & \textit{sofa}  & \textit{lawn}  & overall   \\
    \midrule
    LSeg~\cite{li2022languagedriven}      & 56.0           & 6.0            & 19.2           & 4.5            & 17.5           & 20.6           \\
    ODISE~\cite{xu2023open}     & 52.6           & 24.1           & 52.5           & 48.3           & 39.8           & 43.5           \\
    OV-Seg~\cite{liang2023open}    & 79.8           & 88.9           & 71.4           & 66.1           & 81.2           & 77.5           \\
    \midrule
    FFD~\cite{kobayashi2022decomposing}       & 56.6           & 6.1            & 25.1           & 3.7            & 42.9           & 26.9           \\
    LERF~\cite{kerr2023lerf}      & 73.5           & 53.2           & 46.6           & 27             & 73.7           & 54.8            \\
    3D-OVS~\cite{liu2023weakly}    & 89.5           & 89.3           & 92.8           & 74             & 88.2           & 86.8           \\
    \midrule
    LangSplat   & \textbf{92.5} & \textbf{94.2} & \textbf{94.1} & \textbf{90.0} & \textbf{96.1} & \textbf{93.4} \\ 
    \bottomrule
  \end{tabular}
  \caption{Quantitative comparisons of 3D semantic segmentation on the 3D-OVS dataset. We report the mIoU scores (\%).}
  \label{table:3dovs}
  \vspace{-14pt}
\end{table}

\subsection{Results on the 3D-OVS dataset} 

\textbf{Quantitative Results.} We compare LangSplat with other 2D and 3D state-of-the-art methods on the 3D-OVS dataset in Table \ref{table:3dovs}. We observe that LangSplat not only outperforms 2D-based methods such as ODISE~\cite{xu2023open} and OV-Seg~\cite{liang2023open}, but also achieves better results than 3D-based methods including LERF~\cite{kerr2023lerf} and 3D-OVS~\cite{liu2023weakly} by a large margin. Note that in this dataset, we generate object masks only based on the query text while others, such as 3D-OVS, require the complete category list. In the end, our method achieves an overall mIoU of 93.4\%, 
which shows that LangSplat effectively learns a precise 3D language field.

\noindent\textbf{Qualitative Results.} We present the qualitative results in Figure~\ref{fig:3dovs}. As LERF suffers from the patchy issue and learns over-smoothed features, it fails to find accurate object boundaries. Among all state-of-the-art methods, our methods give the most accurate segmentation maps, which further demonstrates the effectiveness of our LangSplat.

\section{Conclusion}
\label{sec:conclusion}

In this paper, we have presented LangSplat, a method for constructing 3D language fields that enables precise and efficient open-vocabulary querying within 3D spaces. By extending 3D Gaussian Splatting with language features and learning a scene-specific language autoencoder, LangSplat circumvents slow rendering speed associated with NeRF-based methods. Furthermore, we propose to learn the semantic hierarchy defined by SAM, which effectively resolves the point ambiguity problem.
The experimental results clearly demonstrate LangSplat's superiority over existing state-of-the-art methods like LERF, particularly in terms of its remarkable {\speed} $\times$ speed improvement and enhanced performance in open-ended 3D language query tasks.

\noindent \textbf{Acknowledgements} 
This research was funded through National Key Research and Development Program of China (Project No. 2022YFB36066), in part by the Shenzhen Science and Technology Project under Grant (JCYJ20220818101001004, JSGG20210802153150005), and in part by NIH grant R01HD104969, 1U01CA284207, and NSF grant CRCNS-2309041.

{
    \small
    \bibliographystyle{ieeenat_fullname}
    \bibliography{egbib}

\begin{thebibliography}{56}
\providecommand{\natexlab}[1]{#1}
\providecommand{\url}[1]{\texttt{#1}}
\expandafter\ifx\csname urlstyle\endcsname\relax
  \providecommand{\doi}[1]{doi: #1}\else
  \providecommand{\doi}{doi: \begingroup \urlstyle{rm}\Url}\fi

\bibitem[Azuma et~al.(2022)Azuma, Miyanishi, Kurita, and Kawanabe]{azuma2022scanqa}
Daichi Azuma, Taiki Miyanishi, Shuhei Kurita, and Motoaki Kawanabe.
\newblock Scanqa: 3d question answering for spatial scene understanding.
\newblock In \emph{CVPR}, pages 19129--19139, 2022.

\bibitem[Bhalgat et~al.(2023)Bhalgat, Laina, Henriques, Zisserman, and Vedaldi]{bhalgat2023contrastive}
Yash Bhalgat, Iro Laina, Jo{\~a}o~F Henriques, Andrew Zisserman, and Andrea Vedaldi.
\newblock Contrastive lift: 3d object instance segmentation by slow-fast contrastive fusion.
\newblock \emph{arXiv preprint arXiv:2306.04633}, 2023.

\bibitem[Bonvillain(2019)]{bonvillain2019language}
Nancy Bonvillain.
\newblock \emph{Language, culture, and communication: The meaning of messages}.
\newblock Rowman \& Littlefield, 2019.

\bibitem[Caron et~al.(2021)Caron, Touvron, Misra, J{\'e}gou, Mairal, Bojanowski, and Joulin]{caron2021emerging}
Mathilde Caron, Hugo Touvron, Ishan Misra, Herv{\'e} J{\'e}gou, Julien Mairal, Piotr Bojanowski, and Armand Joulin.
\newblock Emerging properties in self-supervised vision transformers.
\newblock In \emph{ICCV}, pages 9650--9660, 2021.

\bibitem[Cascante-Bonilla et~al.(2022)Cascante-Bonilla, Wu, Wang, Feris, and Ordonez]{cascante2022simvqa}
Paola Cascante-Bonilla, Hui Wu, Letao Wang, Rogerio~S Feris, and Vicente Ordonez.
\newblock Simvqa: Exploring simulated environments for visual question answering.
\newblock In \emph{CVPR}, pages 5056--5066, 2022.

\bibitem[Cen et~al.(2023)Cen, Zhou, Fang, Yang, Shen, Xie, Zhang, and Tian]{cen2023segment}
Jiazhong Cen, Zanwei Zhou, Jiemin Fang, Chen Yang, Wei Shen, Lingxi Xie, Xiaopeng Zhang, and Qi Tian.
\newblock Segment anything in 3d with nerfs.
\newblock 2023.

\bibitem[Chan et~al.(2021)Chan, Monteiro, Kellnhofer, Wu, and Wetzstein]{chan2021pi}
Eric~R Chan, Marco Monteiro, Petr Kellnhofer, Jiajun Wu, and Gordon Wetzstein.
\newblock pi-gan: Periodic implicit generative adversarial networks for 3d-aware image synthesis.
\newblock In \emph{CVPR}, pages 5799--5809, 2021.

\bibitem[Chen et~al.(2022)Chen, Xu, Geiger, Yu, and Su]{chen2022tensorf}
Anpei Chen, Zexiang Xu, Andreas Geiger, Jingyi Yu, and Hao Su.
\newblock Tensorf: Tensorial radiance fields.
\newblock In \emph{ECCVn}, pages 333--350. Springer, 2022.

\bibitem[Chen et~al.(2023{\natexlab{a}})Chen, Xia, Ichter, Rao, Gopalakrishnan, Ryoo, Stone, and Kappler]{chen2023open}
Boyuan Chen, Fei Xia, Brian Ichter, Kanishka Rao, Keerthana Gopalakrishnan, Michael~S Ryoo, Austin Stone, and Daniel Kappler.
\newblock Open-vocabulary queryable scene representations for real world planning.
\newblock In \emph{ICRA}, pages 11509--11522. IEEE, 2023{\natexlab{a}}.

\bibitem[Chen et~al.(2023{\natexlab{b}})Chen, Wang, and Liu]{chen2023text}
Zilong Chen, Feng Wang, and Huaping Liu.
\newblock Text-to-3d using gaussian splatting.
\newblock \emph{arXiv preprint arXiv:2309.16585}, 2023{\natexlab{b}}.

\bibitem[Cheng et~al.(2023)Cheng, Li, Xu, Li, Yang, Wang, and Yang]{cheng2023segment}
Yangming Cheng, Liulei Li, Yuanyou Xu, Xiaodi Li, Zongxin Yang, Wenguan Wang, and Yi Yang.
\newblock Segment and track anything.
\newblock \emph{arXiv preprint arXiv:2305.06558}, 2023.

\bibitem[Gao et~al.(2023)Gao, Lin, Xie, Zhou, Cheng, and Yan]{gao2023editanything}
Shanghua Gao, Zhijie Lin, Xingyu Xie, Pan Zhou, Ming-Ming Cheng, and Shuicheng Yan.
\newblock Editanything: Empowering unparalleled flexibility in image editing and generation.
\newblock In \emph{ACM MM, Demo track}, 2023.

\bibitem[Gordon et~al.(2018)Gordon, Kembhavi, Rastegari, Redmon, Fox, and Farhadi]{gordon2018iqa}
Daniel Gordon, Aniruddha Kembhavi, Mohammad Rastegari, Joseph Redmon, Dieter Fox, and Ali Farhadi.
\newblock Iqa: Visual question answering in interactive environments.
\newblock In \emph{CVPR}, pages 4089--4098, 2018.

\bibitem[Huang et~al.(2023{\natexlab{a}})Huang, Mees, Zeng, and Burgard]{huang2023visual}
Chenguang Huang, Oier Mees, Andy Zeng, and Wolfram Burgard.
\newblock Visual language maps for robot navigation.
\newblock In \emph{ICRA}, pages 10608--10615. IEEE, 2023{\natexlab{a}}.

\bibitem[Huang et~al.(2023{\natexlab{b}})Huang, Jiang, Dong, Qiao, Gao, and Li]{huang2023instruct2act}
Siyuan Huang, Zhengkai Jiang, Hao Dong, Yu Qiao, Peng Gao, and Hongsheng Li.
\newblock Instruct2act: Mapping multi-modality instructions to robotic actions with large language model.
\newblock \emph{arXiv preprint arXiv:2305.11176}, 2023{\natexlab{b}}.

\bibitem[Jatavallabhula et~al.(2023)Jatavallabhula, Kuwajerwala, Gu, Omama, Chen, Li, Iyer, Saryazdi, Keetha, Tewari, et~al.]{jatavallabhula2023conceptfusion}
Krishna~Murthy Jatavallabhula, Alihusein Kuwajerwala, Qiao Gu, Mohd Omama, Tao Chen, Shuang Li, Ganesh Iyer, Soroush Saryazdi, Nikhil Keetha, Ayush Tewari, et~al.
\newblock Conceptfusion: Open-set multimodal 3d mapping.
\newblock \emph{arXiv preprint arXiv:2302.07241}, 2023.

\bibitem[Kerbl et~al.(2023)Kerbl, Kopanas, Leimk{\"u}hler, and Drettakis]{kerbl20233d}
Bernhard Kerbl, Georgios Kopanas, Thomas Leimk{\"u}hler, and George Drettakis.
\newblock 3d gaussian splatting for real-time radiance field rendering.
\newblock \emph{TOG}, 42\penalty0 (4):\penalty0 1--14, 2023.

\bibitem[Kerr et~al.(2023)Kerr, Kim, Goldberg, Kanazawa, and Tancik]{kerr2023lerf}
Justin Kerr, Chung~Min Kim, Ken Goldberg, Angjoo Kanazawa, and Matthew Tancik.
\newblock Lerf: Language embedded radiance fields.
\newblock In \emph{ICCV}, pages 19729--19739, 2023.

\bibitem[Kirillov et~al.(2023)Kirillov, Mintun, Ravi, Mao, Rolland, Gustafson, Xiao, Whitehead, Berg, Lo, et~al.]{kirillov2023segment}
Alexander Kirillov, Eric Mintun, Nikhila Ravi, Hanzi Mao, Chloe Rolland, Laura Gustafson, Tete Xiao, Spencer Whitehead, Alexander~C Berg, Wan-Yen Lo, et~al.
\newblock Segment anything.
\newblock In \emph{ICCV}, 2023.

\bibitem[Kobayashi et~al.(2022)Kobayashi, Matsumoto, and Sitzmann]{kobayashi2022decomposing}
Sosuke Kobayashi, Eiichi Matsumoto, and Vincent Sitzmann.
\newblock Decomposing nerf for editing via feature field distillation.
\newblock \emph{NeurIPS}, 35:\penalty0 23311--23330, 2022.

\bibitem[Li et~al.(2022{\natexlab{a}})Li, Weinberger, Belongie, Koltun, and Ranftl]{li2022languagedriven}
Boyi Li, Kilian~Q Weinberger, Serge Belongie, Vladlen Koltun, and Rene Ranftl.
\newblock Language-driven semantic segmentation.
\newblock In \emph{ICLR}, 2022{\natexlab{a}}.

\bibitem[Li et~al.(2022{\natexlab{b}})Li, Huang, Zhu, Tang, Li, Zhou, and Lu]{li2022ordinalclip}
Wanhua Li, Xiaoke Huang, Zheng Zhu, Yansong Tang, Xiu Li, Jie Zhou, and Jiwen Lu.
\newblock Ordinalclip: Learning rank prompts for language-guided ordinal regression.
\newblock \emph{NeurIPS}, 35:\penalty0 35313--35325, 2022{\natexlab{b}}.

\bibitem[Liang et~al.(2023)Liang, Wu, Dai, Li, Zhao, Zhang, Zhang, Vajda, and Marculescu]{liang2023open}
Feng Liang, Bichen Wu, Xiaoliang Dai, Kunpeng Li, Yinan Zhao, Hang Zhang, Peizhao Zhang, Peter Vajda, and Diana Marculescu.
\newblock Open-vocabulary semantic segmentation with mask-adapted clip.
\newblock In \emph{CVPR}, pages 7061--7070, 2023.

\bibitem[Liu et~al.(2023{\natexlab{a}})Liu, Zhan, Zhang, Xu, Yu, Saddik, Theobalt, Xing, and Lu]{liu2023weakly}
Kunhao Liu, Fangneng Zhan, Jiahui Zhang, Muyu Xu, Yingchen Yu, Abdulmotaleb~El Saddik, Christian Theobalt, Eric Xing, and Shijian Lu.
\newblock Weakly supervised 3d open-vocabulary segmentation.
\newblock In \emph{NeurIPS}, 2023{\natexlab{a}}.

\bibitem[Liu et~al.(2023{\natexlab{b}})Liu, Kong, Cen, Chen, Zhang, Pan, Chen, and Liu]{liu2023segment}
Youquan Liu, Lingdong Kong, Jun Cen, Runnan Chen, Wenwei Zhang, Liang Pan, Kai Chen, and Ziwei Liu.
\newblock Segment any point cloud sequences by distilling vision foundation models.
\newblock \emph{arXiv preprint arXiv:2306.09347}, 2023{\natexlab{b}}.

\bibitem[Lu et~al.(2023)Lu, Xiao, Bai, Xiong, and Wang]{lu2023can}
Zhihe Lu, Zeyu Xiao, Jiawang Bai, Zhiwei Xiong, and Xinchao Wang.
\newblock Can sam boost video super-resolution?
\newblock \emph{arXiv preprint arXiv:2305.06524}, 2023.

\bibitem[Luiten et~al.(2023)Luiten, Kopanas, Leibe, and Ramanan]{luiten2023dynamic}
Jonathon Luiten, Georgios Kopanas, Bastian Leibe, and Deva Ramanan.
\newblock Dynamic 3d gaussians: Tracking by persistent dynamic view synthesis.
\newblock \emph{arXiv preprint arXiv:2308.09713}, 2023.

\bibitem[Ma and Wang(2023)]{ma2023segment}
Jun Ma and Bo Wang.
\newblock Segment anything in medical images.
\newblock \emph{arXiv preprint arXiv:2304.12306}, 2023.

\bibitem[Mai et~al.(2023)Mai, Huang, Sun, Song, Mishra, Liu, Gao, Liu, Cong, Hu, et~al.]{mai2023opportunities}
Gengchen Mai, Weiming Huang, Jin Sun, Suhang Song, Deepak Mishra, Ninghao Liu, Song Gao, Tianming Liu, Gao Cong, Yingjie Hu, et~al.
\newblock On the opportunities and challenges of foundation models for geospatial artificial intelligence.
\newblock \emph{arXiv preprint arXiv:2304.06798}, 2023.

\bibitem[Mazurowski et~al.(2023)Mazurowski, Dong, Gu, Yang, Konz, and Zhang]{mazurowski2023segment}
Maciej~A Mazurowski, Haoyu Dong, Hanxue Gu, Jichen Yang, Nicholas Konz, and Yixin Zhang.
\newblock Segment anything model for medical image analysis: an experimental study.
\newblock \emph{Medical Image Analysis}, 89:\penalty0 102918, 2023.

\bibitem[M{\"u}ller et~al.(2022)M{\"u}ller, Evans, Schied, and Keller]{muller2022instant}
Thomas M{\"u}ller, Alex Evans, Christoph Schied, and Alexander Keller.
\newblock Instant neural graphics primitives with a multiresolution hash encoding.
\newblock \emph{TOG}, 41\penalty0 (4):\penalty0 1--15, 2022.

\bibitem[Park et~al.(2021)Park, Sinha, Barron, Bouaziz, Goldman, Seitz, and Martin-Brualla]{park2021nerfies}
Keunhong Park, Utkarsh Sinha, Jonathan~T Barron, Sofien Bouaziz, Dan~B Goldman, Steven~M Seitz, and Ricardo Martin-Brualla.
\newblock Nerfies: Deformable neural radiance fields.
\newblock In \emph{ICCV}, pages 5865--5874, 2021.

\bibitem[Pumarola et~al.(2021)Pumarola, Corona, Pons-Moll, and Moreno-Noguer]{pumarola2021d}
Albert Pumarola, Enric Corona, Gerard Pons-Moll, and Francesc Moreno-Noguer.
\newblock D-nerf: Neural radiance fields for dynamic scenes.
\newblock In \emph{CVPR}, pages 10318--10327, 2021.

\bibitem[Radford et~al.(2021)Radford, Kim, Hallacy, Ramesh, Goh, Agarwal, Sastry, Askell, Mishkin, Clark, et~al.]{radford2021learning}
Alec Radford, Jong~Wook Kim, Chris Hallacy, Aditya Ramesh, Gabriel Goh, Sandhini Agarwal, Girish Sastry, Amanda Askell, Pamela Mishkin, Jack Clark, et~al.
\newblock Learning transferable visual models from natural language supervision.
\newblock In \emph{ICML}, pages 8748--8763. PMLR, 2021.

\bibitem[Shafiullah et~al.(2022)Shafiullah, Paxton, Pinto, Chintala, and Szlam]{shafiullah2022clip}
Nur Muhammad~Mahi Shafiullah, Chris Paxton, Lerrel Pinto, Soumith Chintala, and Arthur Szlam.
\newblock Clip-fields: Weakly supervised semantic fields for robotic memory.
\newblock \emph{arXiv preprint arXiv:2210.05663}, 2022.

\bibitem[Shen et~al.(2023{\natexlab{a}})Shen, Yang, and Wang]{shen2023anything}
Qiuhong Shen, Xingyi Yang, and Xinchao Wang.
\newblock Anything-3d: Towards single-view anything reconstruction in the wild.
\newblock \emph{arXiv preprint arXiv:2304.10261}, 2023{\natexlab{a}}.

\bibitem[Shen et~al.(2022)Shen, Li, Zhu, Duan, Zhou, and Lu]{shen2022learning}
Shuai Shen, Wanhua Li, Zheng Zhu, Yueqi Duan, Jie Zhou, and Jiwen Lu.
\newblock Learning dynamic facial radiance fields for few-shot talking head synthesis.
\newblock In \emph{ECCV}, pages 666--682. Springer, 2022.

\bibitem[Shen et~al.(2023{\natexlab{b}})Shen, Li, Huang, Zhu, Zhou, and Lu]{shen2023sd}
Shuai Shen, Wanhua Li, Xiaoke Huang, Zheng Zhu, Jie Zhou, and Jiwen Lu.
\newblock Sd-nerf: Towards lifelike talking head animation via spatially-adaptive dual-driven nerfs.
\newblock \emph{TMM}, 2023{\natexlab{b}}.

\bibitem[Shen et~al.(2023{\natexlab{c}})Shen, Yang, Yu, Wong, Kaelbling, and Isola]{shen2023distilled}
William Shen, Ge Yang, Alan Yu, Jansen Wong, Leslie~Pack Kaelbling, and Phillip Isola.
\newblock Distilled feature fields enable few-shot language-guided manipulation.
\newblock \emph{arXiv preprint arXiv:2308.07931}, 2023{\natexlab{c}}.

\bibitem[Siddiqui et~al.(2023)Siddiqui, Porzi, Bul{\`o}, M{\"u}ller, Nie{\ss}ner, Dai, and Kontschieder]{siddiqui2023panoptic}
Yawar Siddiqui, Lorenzo Porzi, Samuel~Rota Bul{\`o}, Norman M{\"u}ller, Matthias Nie{\ss}ner, Angela Dai, and Peter Kontschieder.
\newblock Panoptic lifting for 3d scene understanding with neural fields.
\newblock In \emph{CVPR}, pages 9043--9052, 2023.

\bibitem[Sun et~al.(2022)Sun, Sun, and Chen]{sun2022direct}
Cheng Sun, Min Sun, and Hwann-Tzong Chen.
\newblock Direct voxel grid optimization: Super-fast convergence for radiance fields reconstruction.
\newblock In \emph{CVPR}, pages 5459--5469, 2022.

\bibitem[Tang et~al.(2023)Tang, Ren, Zhou, Liu, and Zeng]{tang2023dreamgaussian}
Jiaxiang Tang, Jiawei Ren, Hang Zhou, Ziwei Liu, and Gang Zeng.
\newblock Dreamgaussian: Generative gaussian splatting for efficient 3d content creation.
\newblock \emph{arXiv preprint arXiv:2309.16653}, 2023.

\bibitem[Tschernezki et~al.(2022)Tschernezki, Laina, Larlus, and Vedaldi]{tschernezki2022neural}
Vadim Tschernezki, Iro Laina, Diane Larlus, and Andrea Vedaldi.
\newblock Neural feature fusion fields: 3d distillation of self-supervised 2d image representations.
\newblock In \emph{3DV}, pages 443--453. IEEE, 2022.

\bibitem[Wan et~al.(2024)Wan, Li, Banerjee, Adhinarta, Sjostedt, Wu, Lichtman, Pfister, and Wei]{wan2024trisam}
Jia Wan, Wanhua Li, Atmadeep Banerjee, Jason~Ken Adhinarta, Evelina Sjostedt, Jingpeng Wu, Jeff Lichtman, Hanspeter Pfister, and Donglai Wei.
\newblock Trisam: Tri-plane sam for zero-shot cortical blood vessel segmentation in vem images.
\newblock \emph{arXiv preprint arXiv:2401.13961}, 2024.

\bibitem[Wu et~al.(2023)Wu, Yi, Fang, Xie, Zhang, Wei, Liu, Tian, and Wang]{wu20234d}
Guanjun Wu, Taoran Yi, Jiemin Fang, Lingxi Xie, Xiaopeng Zhang, Wei Wei, Wenyu Liu, Qi Tian, and Xinggang Wang.
\newblock 4d gaussian splatting for real-time dynamic scene rendering.
\newblock \emph{arXiv preprint arXiv:2310.08528}, 2023.

\bibitem[Xu et~al.(2023)Xu, Liu, Vahdat, Byeon, Wang, and De~Mello]{xu2023open}
Jiarui Xu, Sifei Liu, Arash Vahdat, Wonmin Byeon, Xiaolong Wang, and Shalini De~Mello.
\newblock Open-vocabulary panoptic segmentation with text-to-image diffusion models.
\newblock In \emph{CVPR}, pages 2955--2966, 2023.

\bibitem[Yang et~al.(2023{\natexlab{a}})Yang, Gao, Li, Gao, Wang, and Zheng]{yang2023track}
Jinyu Yang, Mingqi Gao, Zhe Li, Shang Gao, Fangjing Wang, and Feng Zheng.
\newblock Track anything: Segment anything meets videos.
\newblock \emph{arXiv preprint arXiv:2304.11968}, 2023{\natexlab{a}}.

\bibitem[Yang et~al.(2023{\natexlab{b}})Yang, Gao, Zhou, Jiao, Zhang, and Jin]{yang2023deformable}
Ziyi Yang, Xinyu Gao, Wen Zhou, Shaohui Jiao, Yuqing Zhang, and Xiaogang Jin.
\newblock Deformable 3d gaussians for high-fidelity monocular dynamic scene reconstruction.
\newblock \emph{arXiv preprint arXiv:2309.13101}, 2023{\natexlab{b}}.

\bibitem[Yang et~al.(2023{\natexlab{c}})Yang, Yang, Pan, Zhu, and Zhang]{yang2023real}
Zeyu Yang, Hongye Yang, Zijie Pan, Xiatian Zhu, and Li Zhang.
\newblock Real-time photorealistic dynamic scene representation and rendering with 4d gaussian splatting.
\newblock \emph{arXiv preprint arXiv:2310.10642}, 2023{\natexlab{c}}.

\bibitem[Yao et~al.(2023)Yao, Wang, Ye, and Liu]{yao2023matte}
Jingfeng Yao, Xinggang Wang, Lang Ye, and Wenyu Liu.
\newblock Matte anything: Interactive natural image matting with segment anything models.
\newblock \emph{arXiv preprint arXiv:2306.04121}, 2023.

\bibitem[Yi et~al.(2023)Yi, Fang, Wu, Xie, Zhang, Liu, Tian, and Wang]{yi2023gaussiandreamer}
Taoran Yi, Jiemin Fang, Guanjun Wu, Lingxi Xie, Xiaopeng Zhang, Wenyu Liu, Qi Tian, and Xinggang Wang.
\newblock Gaussiandreamer: Fast generation from text to 3d gaussian splatting with point cloud priors.
\newblock \emph{arXiv preprint arXiv:2310.08529}, 2023.

\bibitem[Yu et~al.(2021)Yu, Ye, Tancik, and Kanazawa]{yu2021pixelnerf}
Alex Yu, Vickie Ye, Matthew Tancik, and Angjoo Kanazawa.
\newblock pixelnerf: Neural radiance fields from one or few images.
\newblock In \emph{CVPR}, pages 4578--4587, 2021.

\bibitem[Yu et~al.(2023)Yu, Feng, Feng, Liu, Jin, Zeng, and Chen]{yu2023inpaint}
Tao Yu, Runseng Feng, Ruoyu Feng, Jinming Liu, Xin Jin, Wenjun Zeng, and Zhibo Chen.
\newblock Inpaint anything: Segment anything meets image inpainting.
\newblock \emph{arXiv preprint arXiv:2304.06790}, 2023.

\bibitem[Zhi et~al.(2021)Zhi, Laidlow, Leutenegger, and Davison]{zhi2021place}
Shuaifeng Zhi, Tristan Laidlow, Stefan Leutenegger, and Andrew~J Davison.
\newblock In-place scene labelling and understanding with implicit scene representation.
\newblock In \emph{ICCV}, pages 15838--15847, 2021.

\bibitem[Zhou et~al.(2022)Zhou, Yang, Loy, and Liu]{zhou2022learning}
Kaiyang Zhou, Jingkang Yang, Chen~Change Loy, and Ziwei Liu.
\newblock Learning to prompt for vision-language models.
\newblock \emph{IJCV}, 130\penalty0 (9):\penalty0 2337--2348, 2022.

\bibitem[Zwicker et~al.(2001)Zwicker, Pfister, Van~Baar, and Gross]{zwicker2001ewa}
Matthias Zwicker, Hanspeter Pfister, Jeroen Van~Baar, and Markus Gross.
\newblock Ewa volume splatting.
\newblock In \emph{VIS}, pages 29--538. IEEE, 2001.

\end{thebibliography}
}

\clearpage
\appendix
\maketitlesupplementary

\section{Video Demo}
In Figure 1 of our main paper, we have visualized the language features learned by LERF and our method. For a fair comparison, we perform PCA on the decoded feature $\Psi(\bm{F_t^l}) \in \mathbb{R}^{D \times H \times W}$ for our method. However, one benefit of our method is that we are able to directly visualize the learned language features in the encoded 3-dimensional latent space, which can ensure color consistency between frames.\footnote{The consistency of color in PCA visualizations across different frames is not ensured.} Specifically, we normalize the encoded 3-dimensional latent features $\bm{H}_t^l(v)  \in \mathbb{R}^{3 \times H \times W}$ and visualize them by treating the 3-dimensional features as RGB channels.

We strongly recommend readers refer to our video demo at \url{https://www.youtube.com/watch?v=XMlyjsei-Es} to observe the learned 3D language fields in the scene-specific latent space. The video demonstrates that our method has acquired a 3D language representation that is both 3D-consistent and distinctly shaped, which significantly distinguishes it from existing methods that often only learn 3D language representations with blurred boundaries. Meanwhile, our approach achieves a speedup of 119 $\times$ compared to LERF at a resolution of $988 \times 731$ and further improves to 199 $\times$ faster at a resolution of $1440 \times 1080$.

\section{More Implementation Details}
For each text query, we can obtain three relevancy maps with our trained 3D language Gaussians, each representing one semantic level defined by SAM. Then we use different strategies to choose the best semantic level and obtain the predictions for different tasks.  

\noindent\textbf{3D Object Localization on LERF.} 
To mitigate the impact of outliers, we first employ a mean convolution filter with a size of 20 to smooth the values of three relevancy maps. For the smoothed relevancy maps, we select the one with the highest smoothed relevancy score and take the corresponding position as the final prediction.

\noindent\textbf{3D Semantic Segmentation on LERF.}
Similarly, to mitigate the influence of outliers, we apply a mean filter with a size of 20 to smooth the three relevancy maps. Subsequently, we select the relevancy map with the maximum smoothed relevancy score for binary mask prediction.
For the selected relevancy map, we first normalize its relevancy scores and then use a threshold to obtain a binary image as the final prediction mask.

\noindent\textbf{3D Semantic Segmentation on 3D-OVS.} 
For each class query, we obtain three relevancy maps. We apply a threshold of 0.4 to these relevancy maps, setting relevancy scores below 0.4 to 0 and relevancy scores above 0.4 to 1, resulting in three binary maps. We calculate the average relevancy scores within the mask region for each relevancy map and select the relevancy map with the highest average response as the final predicted binary map.

\section{More Quantitative Results}
In addition to the mIoU metric, the Accuracy metric is also employed on the 3D-OVS dataset in~\cite{liu2023weakly}.\footnote{After checking with the authors of 3D-OVS, we confirmed that the mAP results reported in~\cite{liu2023weakly} are, in fact, the Accuracy results. } Therefore, we also compare our method with other state-of-the-art methods on the 3D-OVS dataset using the Accuracy metric. The results are shown in Table \ref{table:acc3dovs}. We observe that our method consistently outperforms other methods, which further illustrates the superiority of our method.

\begin{table}[t]
  \renewcommand\tabcolsep{4pt}
  \centering
  \begin{tabular}{lcccccc}
    \toprule
    Method & \textit{bed}   & \textit{bench} & \textit{room}  & \textit{sofa}  & \textit{lawn}  & overall   \\
    \midrule
    LSeg~\cite{li2022languagedriven}      & 87.6           & 42.7            & 46.1           & 16.5            & 77.5           & 54.1           \\
    ODISE~\cite{xu2023open}     & 86.5           & 39.0           & 59.7           & 35.4           & 82.5           & 60.6           \\
    OV-Seg~\cite{liang2023open}    & 40.4           & 89.2           & 49.1           & 69.6           & 92.1           & 68.1           \\
    \midrule
    FFD~\cite{kobayashi2022decomposing}       & 86.9           & 42.8            & 51.4           & 9.5            & 82.6           & 54.6           \\
    LERF~\cite{kerr2023lerf}      & 86.9           & 79.7           & 79.8           & 43.8             & 93.5           & 76.7            \\
    3D-OVS~\cite{liu2023weakly}    & 96.7           & 96.3           & 98.9           & 91.6             & 97.3           & 96.2           \\
    \midrule
    LangSplat   & \textbf{99.2} & \textbf{98.6} & \textbf{99.3} & \textbf{97.9} & \textbf{99.4} & \textbf{98.9} \\ 
    \bottomrule
  \end{tabular}
  \caption{Quantitative comparisons of 3D semantic segmentation on the 3D-OVS dataset. We report the accuracy scores (\%).}
  \label{table:acc3dovs}
\end{table}

\section{More Ablation Study}

\begin{table}[t]
  \renewcommand\tabcolsep{5pt}
  \centering
  \begin{tabular}{lcccc}
    \toprule
    $d$ & 1 & 2 & 3  & 8 \\
    \midrule
   mIoU (\%) & 6.46 & 91.93 & 94.19  & 95.20 \\
   Speed (s/q) & 0.2770 & 0.2779 & 0.2788  & 0.2807 \\
    \bottomrule
  \end{tabular}
  \caption{The ablations of latent dimension $d$ for our scene-specific autoencoder. The results are obtained on the bench scene of the 3D-OVS dataset. The image resolution is $1440 \times 1080$.}
  \label{table:ablationLatent}
\end{table}

To reduce the memory cost of our 3D language Gaussians, we proposed the scene-specific autoencoder to learn a latent feature. We show the ablation results of different latent dimensions $d$ on the bench scene of the 3D-OVS dataset in Table \ref{table:ablationLatent}. We observed that as $d$ increases, the mIoU performance improves, with only a slight increase in the time cost. We chose $d=3$ because it allows us to directly visualize the learned 3D language field in the latent space by treating the 3-dimensional features as the RGB channels. We also strongly encourage readers to refer to our video demo to observe how our learned language field accurately captures the precise 3D shape of objects in the scene-specific latent space.

\begin{figure*}[t]
  \centering
   \includegraphics[width=1\linewidth]{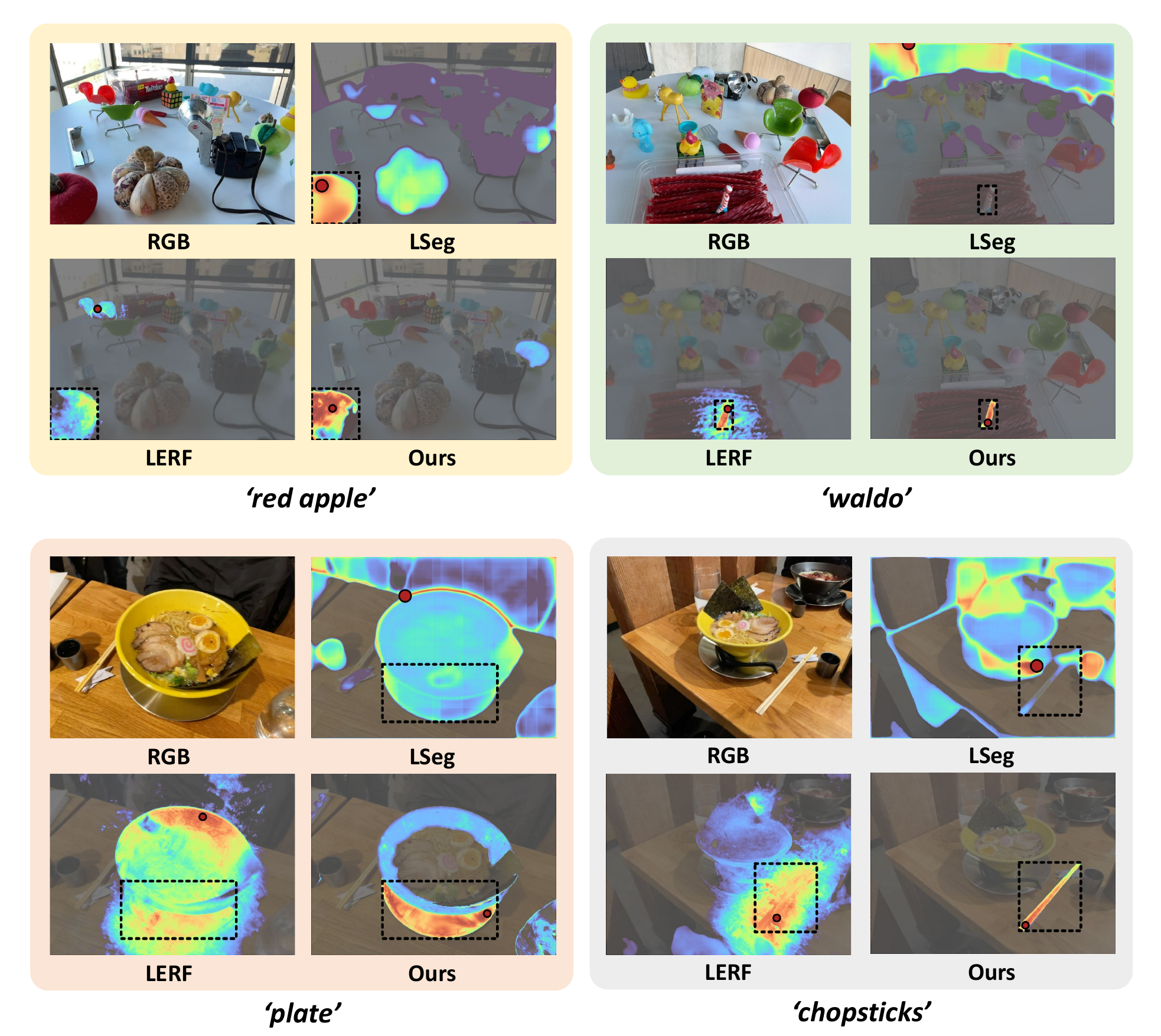}
   \caption{More qualitative comparisons of open-vocabulary 3D object localization on the LERF dataset. The red points are the model predictions and the black dashed bounding boxes denote the annotations.}
   \label{fig:morelerfloc}
\end{figure*}

\section{More Visualization Results}

\noindent\textbf{3D Object Localization on LERF.} We visualize more examples on the LERF dataset for open-vocabulary 3D object localization in Figure \ref{fig:morelerfloc}. We found that for text queries such as ``red apple" and ``plate", LERF failed to correctly locate the 3D positions, whereas our method succeeded. For text queries like ``waldo" and ``chopsticks", although LERF could identify the correct location, its activation values were more dispersed, whereas our method was able to focus more precisely on the queried object.

\begin{figure*}[t]
  \centering
   \includegraphics[width=1\linewidth]{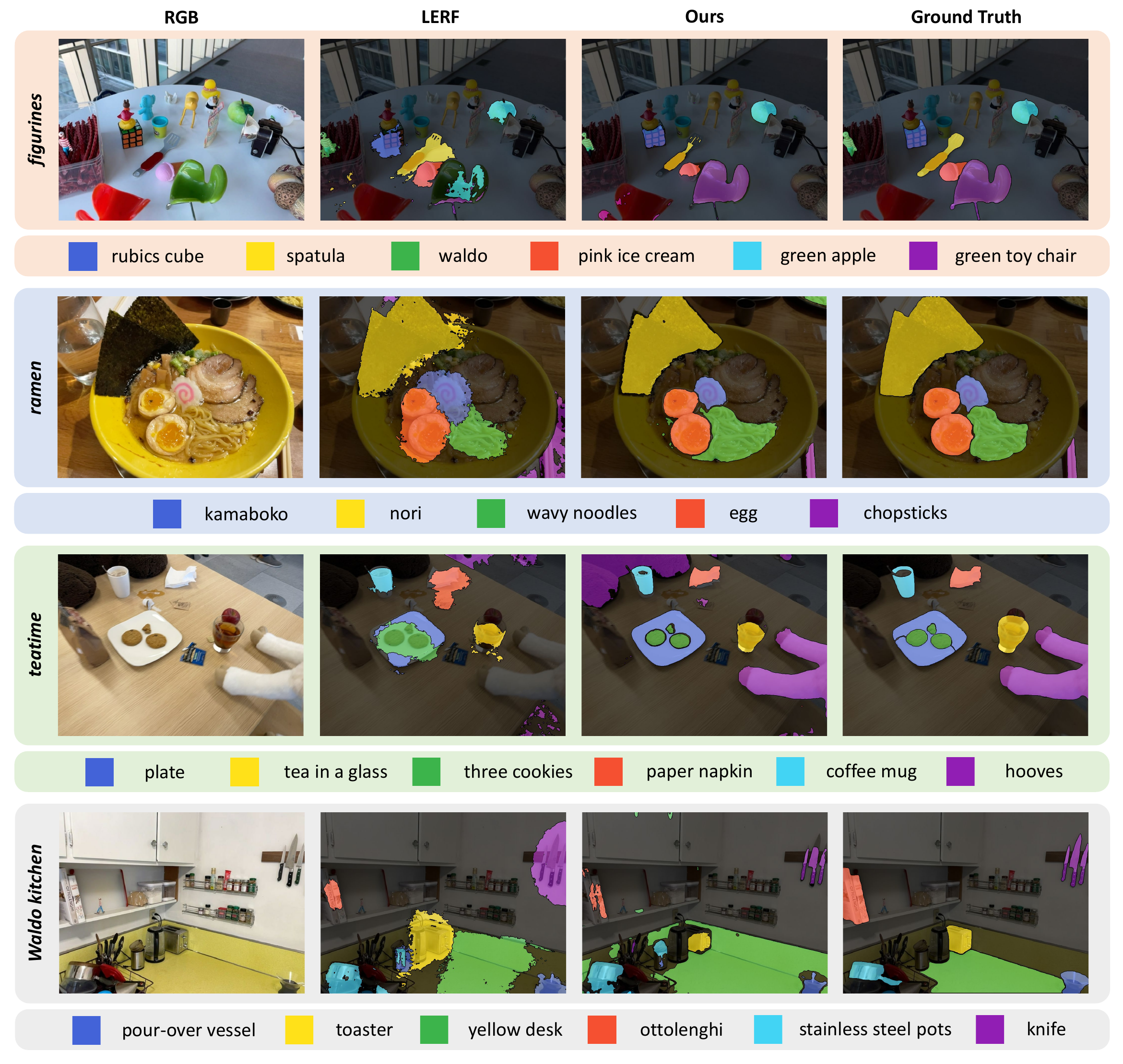}
   \caption{More qualitative comparisons of open-vocabulary 3D semantic segmentation on the LERF dataset.}
   \label{fig:morelerfseg}
\end{figure*}

\noindent\textbf{3D Semantic Segmentation on LERF.} We demonstrate more examples on the LERF dataset for open-vocabulary 3D semantic segmentation in Figure \ref{fig:morelerfseg}. We observed that the results produced by LERF were unable to provide the precise shape of the queried object and exhibited a significant amount of noise, whereas our method could accurately depict the object's shape. These results show the effectiveness of our proposed LangSplat.

\begin{figure*}[ht]
  \centering
   \includegraphics[width=1\linewidth]{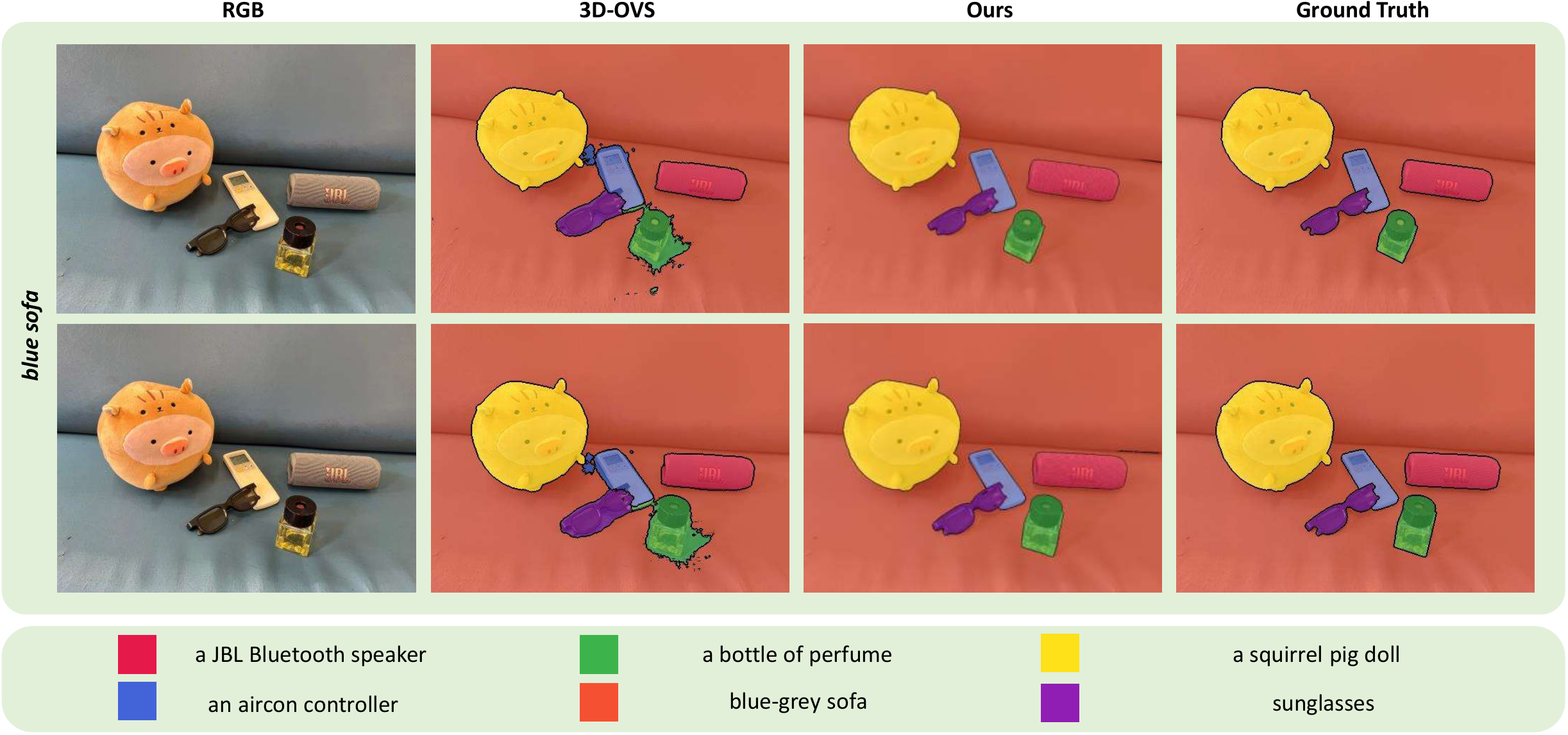}
   \caption{Qualitative comparisons on the blue sofa scene of the 3D-OVS dataset.}
   \label{fig:bluesofa}
\end{figure*}

\begin{figure*}[ht]
  \centering
   \includegraphics[width=1\linewidth]{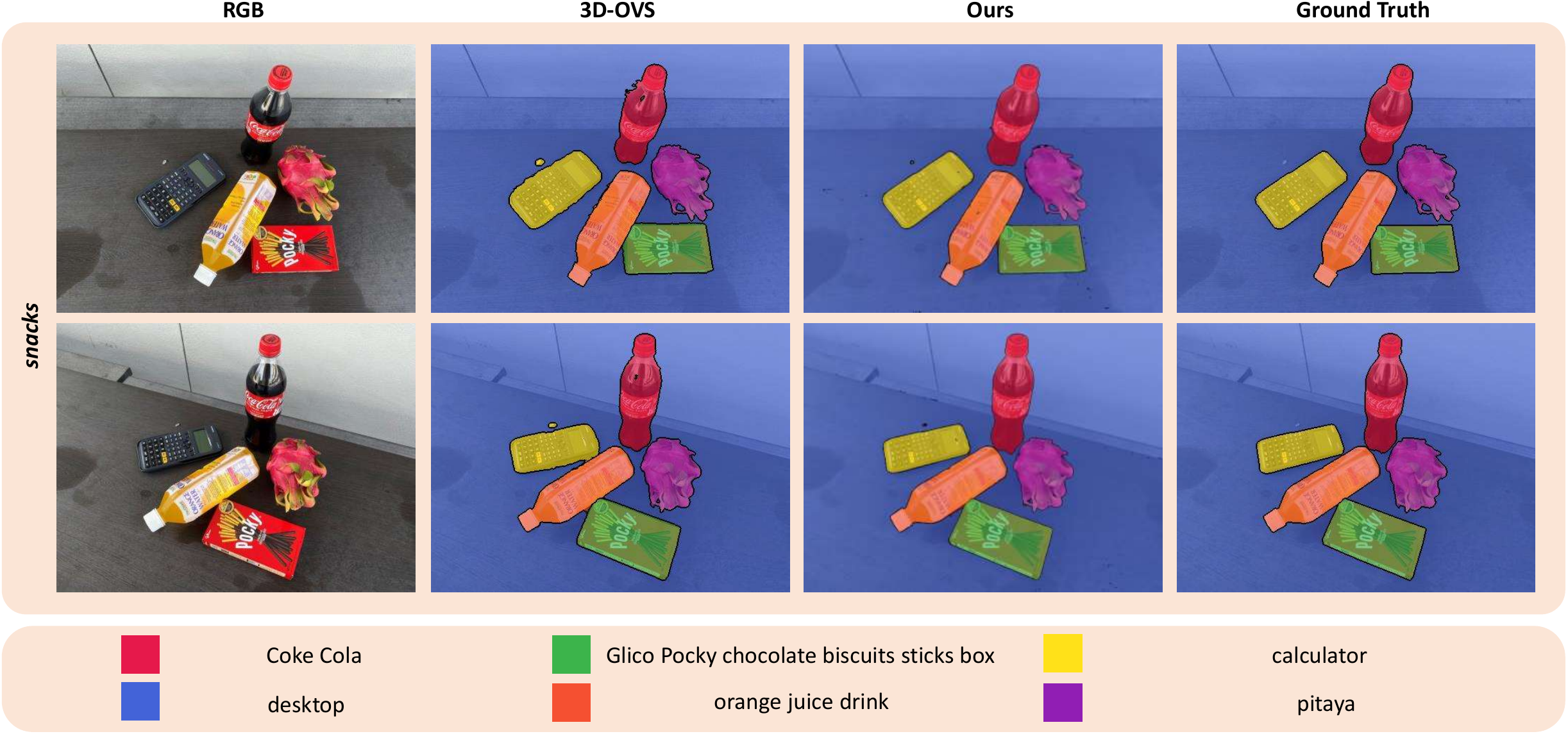}
   \caption{Qualitative comparisons on the snacks scene of the 3D-OVS dataset.}
   \label{fig:snacks}
\end{figure*}

\begin{figure*}[ht]
  \centering
   \includegraphics[width=1\linewidth]{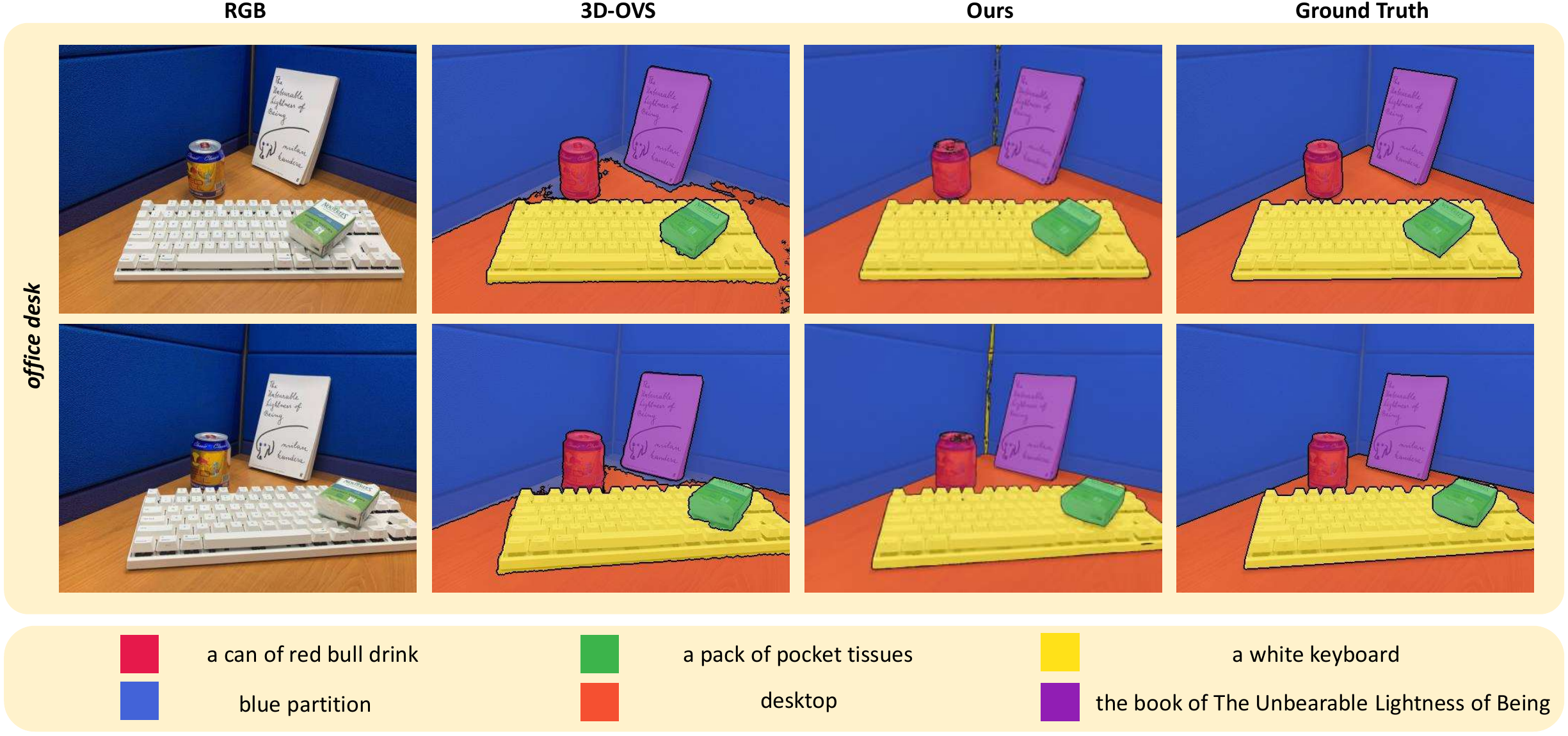}
   \caption{Qualitative comparisons on the office desk scene of the 3D-OVS dataset.}
   \label{fig:officedesk}
\end{figure*}

\begin{figure*}[ht]
  \centering
   \includegraphics[width=1\linewidth]{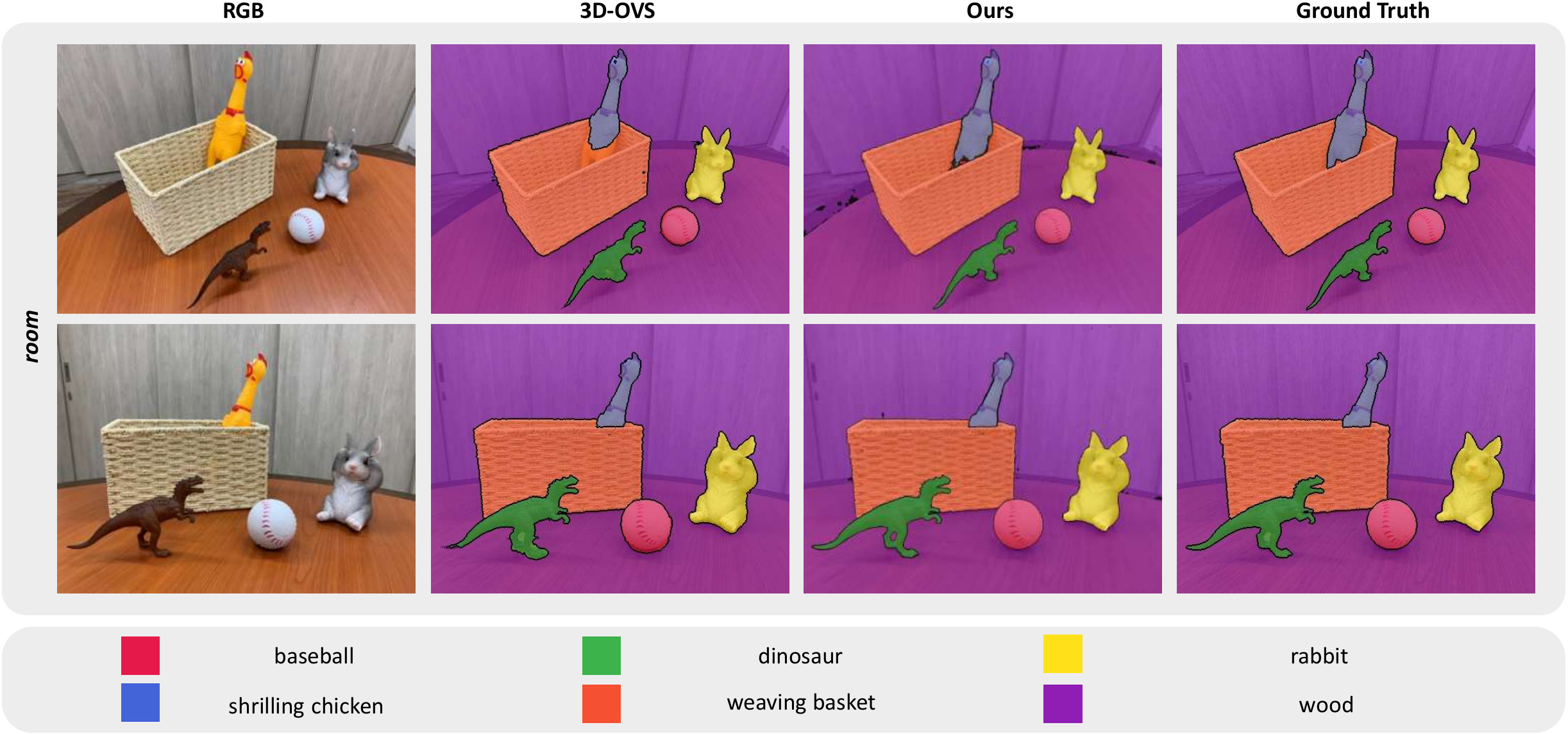}
   \caption{Qualitative comparisons on the room scene of the 3D-OVS dataset..}
   \label{fig:room}
\end{figure*}

\noindent\textbf{3D Semantic Segmentation on 3D-OVS.} We show more scenes on the 3D-OVS dataset for open-vocabulary 3D semantic segmentation in Figures \ref{fig:bluesofa}, \ref{fig:snacks}, \ref{fig:officedesk}, and \ref{fig:room}, respectively. Compared to the previous state-of-the-art method 3D-OVS, our approach provides more precise object boundaries and exhibits reduced noise, which illustrates that our LangSplat learns a more accurate 3D language field.

\end{document}